\documentclass[letterpaper,12pt]{article}

\usepackage[utf8]{inputenc}
\usepackage[T1]{fontenc}
\usepackage[english]{babel}
\usepackage{mathtools}
\usepackage{mleftright}
\usepackage[mathscr]{euscript}
\usepackage{dutchcal}
\usepackage{graphicx,url}
\usepackage[left=1in,right=1in,top=1in,bottom=1in,letterpaper]{geometry}

\usepackage{dsfont}
\usepackage{setspace}
\usepackage{amssymb}
\usepackage[margin = 1cm]{caption}
\usepackage{sidecap}
\usepackage{float}
\usepackage{amsmath}
\usepackage{amsthm}
\usepackage{color,epstopdf}
\usepackage[noend]{algpseudocode}
\usepackage{hyperref}
\usepackage{algorithm,algpseudocode}
\usepackage{breqn}

\usepackage{epsfig}
\usepackage{tabu}
\usepackage[table,xcdraw]{xcolor}
\usepackage{arydshln}
\usepackage{xcolor}
\usepackage{mathrsfs}
\usepackage{caption}
\usepackage{subcaption}

\DeclarePairedDelimiterX{\expectarg}[1]{[}{]}{%
  \ifnum\currentgrouptype=16 \else\begingroup\fi
  \activatebar#1
  \ifnum\currentgrouptype=16 \else\endgroup\fi
}

\newcommand{\LinesNumbered}{
  \setboolean{algocf@linesnumbered}{true}%
  \renewcommand{\algocf@linesnumbered}{\everypar={\nl}}}%
\let\oldnl\nl
\newcommand{\nonl}{\renewcommand{\nl}{\let\nl\oldnl}}

\newcommand{\innermid}{\nonscript\;\delimsize\vert\nonscript\;}
\newcommand\tab[1][1cm]{\hspace*{#1}}

\newcommand{\activatebar}{%
  \begingroup\lccode`\~=`\|
  \lowercase{\endgroup\let~}\innermid 
  \mathcode`|=\string"8000
}

\algnewcommand{\Inputs}[1]{%
  \State \textbf{Inputs:}
  \Statex \hspace*{\algorithmicindent}\parbox[t]{.8\linewidth}{\raggedright #1}
}
\algnewcommand{\Initialize}[1]{%
  \State \textbf{Initialize:}
  \Statex \hspace*{\algorithmicindent}\parbox[t]{.8\linewidth}{\raggedright #1}
}
\algnewcommand{\Data}[1]{%
  \State \textbf{Data:}
  \Statex \hspace*{\algorithmicindent}\parbox[t]{.8\linewidth}{\raggedright #1}
}

\DeclareMathOperator*{\argmin}{\arg\!\min}
\newtheorem{theorem}{Theorem}

\graphicspath{{images/}}

\begin{document}
\title{Tricks and Plugins to GBM on Images and Sequences }

\author{
  Biyi Fang\\
  Department of Engineering Science and Applied Mathematics\\
  Northwestern University\\
  \texttt{biyifang2021@u.northwestern.edu} \\
  \and
  Jean Utke\\
  Data Discovery \& Decision Science \\
  Allstate\\
  \texttt{jutke@allstate.com}\\
  \and
  Diego Klabjan \\
  Department of Industrial Engineering and Management Sciences \\
  Northwestern University\\
  \texttt{d-klabjan@northwestern.edu} \\
}

\maketitle

\begin{abstract}
\noindent Convolutional neural networks (CNNs) and transformers, which are composed of multiple processing layers and blocks to learn the representations of data with multiple abstract levels, are the most successful machine learning models in recent years. However, millions of parameters and many blocks make them difficult to be trained, and sometimes several days or weeks are required to find an ideal architecture or tune the parameters. Within this paper, we propose a new algorithm for boosting Deep Convolutional Neural Networks (BoostCNN) to combine the merits of dynamic feature selection and BoostCNN, and another new family of algorithms combining boosting and transformers. To learn these new models, we introduce subgrid selection and importance sampling strategies and propose a set of algorithms to incorporate boosting weights into a deep learning architecture based on a least squares objective function. These algorithms not only reduce the required manual effort for finding an appropriate network architecture but also result in superior performance and lower running time. Experiments show that the proposed methods outperform benchmarks on several fine-grained classification tasks.\\
\end{abstract}

\section{Introduction}
\tab Deep convolutional neural networks (CNNs) and transformers such as BERT have had great recent success in learning image representations for vision tasks and NLP, respectively. Given the outstanding results produced by these networks, they have been widely applied in image classification (\cite{He2016DeepRL}, \cite{Krizhevsky2017ImageNetCW}, \cite{Lin2015BilinearCM}), object detection (\cite{Girshick2014RichFH}, \cite{Iandola2014DenseNetIE}, \cite{Ren2015FasterRT}), speech recognition, (\cite{Karita2019ImprovingTE}, \cite{Yeh2019TransformerTransducerES}, \cite{Dong2018SpeechTransformerAN}), and language translation (\cite{Li2019NeuralSS}, \cite{Gangi2019AdaptingTT}, \cite{Zhou2018SyllableBasedSS}). However, an optimal image or text representation for each task is unique and finding an optimal deep neural network structure is a challenging problem. There are some approaches (neural architecture search) for designing these deep networks such as AutoML for Model Compression (AMC) in \cite{He2018AMCAF} and LEAF in \cite{Liang2019EvolutionaryNA}, however, these methods require weeks of training on thousands of GPUs. Meanwhile, ensemble methods for classification and regression have gained a lot of attention in recent years, which perform, both theoretically and empirically, substantially better than single models in a wide range of tasks, i.e. boosting decision trees \cite{Quinlan2004InductionOD}. In order to tackle the design challenge specifically for CNNs, an idea of combining boosting and shallow CNNs is proposed in \cite{Moghimi2016BoostedCN}. Their idea is to simplify the complicated design process of deep neural networks by employing the boosting strategy which combines the strengths of multiple CNNs. However, the memory requirement and running time become challenging when the weak learner is not extremely simple. Moreover, very limited contribution has been made to the case when the weak learner is a transformer. Furthermore, no work has been conducted around the idea of only using partial data with weak learners.

In this paper, we propose a family of boosting algorithms for images, namely subgrid BoostCNN, and another family of boosting algorithms for sequences, namely BoostTransformer, which are both based on boosting, deep CNNs and transformers. We select a subset of features for each weak learner, where the concepts are borrowed from random forests. This strategy requires new ideas in order to accommodate unstructured data. Moreover, we apply the concept of importance sampling, which assigns a probability to each sample, to the combination of boosting and a transformer. 

Subgrid BoostCNN aims to solve the same problem as deep CNNs but provides higher accuracy with lower running time and memory requirements. Subgrid BoostCNN builds on the previous boosting Deep Convolutional Neural Networks (BoostCNN) \cite{Moghimi2016BoostedCN}. One important new aspect in subgrid BoostCNN is that it does not require a full image for training a weak learner; instead, it  selects only important pixels based on the gradient from each image combined with the corresponding residual to train the current weak learner. This implies  breaking the original relationship between a pixel and its neighborhood, thus possibly leading to noisier training, but the subgrid BoostCNN does reduce computation to important pixels. Another option to reduce the running time is to skip the optimization process for the original full CNN, which finds the important pixels for the weak learner. Instead, we borrow the CNN portion from the last weak learner concatenated with the fully connected layer used in the first iterate to compute the importance value of each pixel, and train the CNN concatenated with an appropriate fully connected layer. Consequently, subgrid BoostCNN does the optimization process once in each iteration, which is the same as BoostCNN, while subgrid BoostCNN has fewer parameters when compared with BoostCNN. This subgrid trick is essential especially when the training process for the weak learner is computationally demanding. Furthermore, we demonstrate subgrid BoostCNN on three different image datasets and argue that subgrid BoostCNN outperforms both BoostCNN and deep CNNs. More precisely, subgrid BoostCNN improves the accuracies by $1.16\%, 0.82\%, 12.10\%$ on CIFAR-10, SVHN and ImageNetSub datasets, respectively, when compared to standard CNN models. In addition, subgrid BoostCNN obtains accuracies $0.34\%, 0.50\%, 4.19\%$ higher than those generated by BoostCNN on the aforementioned datasets, respectively.

BoostTransformer is an algorithm which combines the merits of boosting and transformers. BoostTransformer incorporates boosting weights with transformers based on least squares objective functions. Motivated by the successful combination of BoostCNN and the subgrid trick, we propose subsequence BoostTransformer, which does not require the full data for training weak learners. In subsequence BoostTransformer, important tokens, which are from the input, are selected for each weak learner based on the attention distribution \cite{Vaswani2017AttentionIA}. Similarly, we might lose the direct connections between consecutive words, while informative words are retained for learning. Consequently, subsequence BoostTransformer takes less time to achieve a better accuracy when compared to vanilla BoostTransformer. Moreover, motivated by the phenomenon that overfitting in BoostTransformer appears early, we propose a new algorithm, namely importance-sampling-based BoostTransformer, which combines the merits of BoostTransformer and importance sampling. Importance-sampling-based BoostTransformer first computes a probability distribution for all the samples in the dataset; then in each iteration, it randomly chooses a subset of samples based on the pre-computed probability distribution; lastly, similar to BoostTransformer, it trains the weak learner on the selected samples. This algorithm not only delays overfitting, but also improves the accuracy and significantly reduces the running time. We present a complete technical proof for importance-sampling-based BoostTransformer showing that the optimal probability distribution is proportional to the norm of the residuals. Lastly, we conduct computational experiments demonstrating a superior performance of the proposed algorithms. More precisely, BoostTransformer provides higher accuracy and more stable solutions when compared to transformers. Moreover, subsequence BoostTransformer and importance-sampling-based BoostTransformer not only provide better and more robust solutions but also dramatically reduce the running time when compared to transformers. Compared to standard transformers, BoostTransformer, subsequence BoostTransformer and importance-sampling-based BoostTransformer provide an average of $0.87\%, 0.55\%, 0.79\%$ accuracy improvements, respectively, on IMDB, Yelp and Amazon datasets. Furthermore, subsequence BoostTransformer and importance-sampling-based BoostTransformer take only two thirds and one half of time transformers need to learn the datasets, respectively.

In summary, we make the following contributions.
\begin{itemize}
    \item We provide a better boosting method for deep CNNs, i.e. subgrid BoostCNN, which requires only important pixels from the image dataset where such pixels are selected dynamically for each weak learner.
    \item We provide a boosting method for sequences, i.e. BoostTransformer, which combines the merits of boosting and transformers.
    \item We provide a better boosting method for transformers, i.e. subsequence BoostTransformer, which does not require the full sequences but only important tokens.
    \item We provide another enhancement for BoostTransformer, i.e. importance-sampling-based BoostTransformer, which combines importance sampling and BoostTranformer. Moreover, we provide a proof showing that the optimal probability distribution for the samples is proportional to the norm of the residuals.
    \item We present numerical results showing that subsequence BoostTransformer and importance-sampling-based BoostTransformer outperform vanilla transformers on select tasks and datasets.
\end{itemize}
The rest of the paper is organized as follows. In the next section, we review several related works in gradient boosting machine, CNN and transformers. In Section 3, we state the formal optimization problem and provide the exposition of the subgrid BoostCNN. In the subsequent section, we propose BoostTransformer, subsequence BoostTransformer and importance-sampling-based BoostTransformer, followed by the analysis of the optimal probability distribution for importance-sampling-based BoostTransformer. In Section 5, we present experimental results comparing the different algorithms. 

\section{Related Work}
There are many extensions of Gradient Boosting Machine (GBM) \cite{Natekin2013GradientBM}, however, a full retrospection of this immense literature exceeds the scope of this work. In this section, we mainly state several kinds of variations of GBM which are most related to our new algorithms, together with the two add-ons to our optimization algorithms, i.e. subgrid and importance sampling.

\textit{\textbf{Boosting for CNNs:}}
Deep CNNs, which have recently produced outstanding performance in learning image representations, are capable of learning complex features that are highly invariant and discriminant \cite{Gu2018RecentAI}. The success of deep CNNs in recognizing objects has encouraged recent works to combine boosting together with deep CNNs. Brahimi $\&$ Aoun \cite{Brahimi2019BoostedCN} propose a new Boosted Convolutional Neural Network architecture, which uses a very deep convolutional neural network reinforced by adding Boosted Blocks which consist of a succession of convolutional layers boosted by using a Multi-Bias Nonlinear Activation function.  Nevertheless, the architecture of the proposed Boosted convolutional neural network is fixed; it can not dynamically change the number of Boosted Blocks for a given dataset. Another attempt at combining deep CNNs and boosting is boosted sampling \cite{Berger2018BoostedTO}, which uses posterior error maps, generated throughout training, to focus sampling on different regions, resulting in a more informative loss. However, boosted sampling applies boosting on selected samples and treats deep CNN as a black box to make a prediction. To enrich the usage of the information generated by deep CNNs, Lee $\&$ Chen \cite{Lee2018ImageCB} propose a new BoostCNN structure which employs a trained deep convolutional neural network model to extract the features of the images and then use the AdaBoost algorithm to assemble the Softmax classifiers. However, it remains unclear how to combine different sets of the features extracted and the computational cost is high when training several deep CNNs at the same time. To tackle this problem, Han $\&$ Meng \cite{Han2016IncrementalBC} propose Incremental Boosting CNN (IB-CNN) to integrate boosting into the CNN via an incremental boosting layer that selects discriminative neurons from a lower layer and is incrementally updated on successive mini-batches. Different from IB-CNN which only involves one deep CNN, BoostCNN \cite{Moghimi2016BoostedCN} incorporates boosting weights into the neural network architecture based on least squares objective functions, which leads to the aggregation of several CNNs. However, the computational and memory demand of BoostCNN is high when the weak learner is not simple. All of the above train the weak learners on all features.

\textit{\textbf{Boosting for Recurrent Neural Network (RNN) and Transformer:}}
RNN, long short-term memory (LSTM) and transformers have been firmly established as state of the art approaches in sequence modeling and transduction problems such as language modeling and machine translation (\cite{Bahdanau2015NeuralMT}, \cite{Cho2014LearningPR}, \cite{Sutskever2014SequenceTS}, \cite{Vaswani2017AttentionIA}). Some efforts have been made to combine boosting with RNN or LSTM. Chen $\&$ Lundberg \cite{Chen2018HybridGB} present feature learning via LSTM networks and prediction via gradient boosting trees (XGB). More precisely, they generate features by performing supervised representation learning with an LSTM network, then augment the original XGB model with these new generated features. However, the selection of the features from LSTM is not determined by XGB, which leads to a disconnect between LSTM and XGB. Another attempt at combining boosting and RNN is the boosting algorithm for regression with RNNs \cite{Assaad2008ANB}. This algorithm adapts an ensemble method to the problem of predicting future values of time series using RNNs as base learners, and it is based on the boosting algorithm where different points of the time series are emphasized during the learning process by training different base learners on different subsets of time points. However, combing boosting and transformers has not previously been investigated . Although, analyses of attention in Transformer have been explored \cite{Clark2019WhatDB}, using  the attention distribution in token selection has not been extensively studied.

Given the fact that importance sampling improves the performance by prioritizing training samples, importance sampling has been well studied, both theoretically and empirically, in standard stochastic gradient descent settings \cite{Needell2014StochasticGD} \cite{Zhao2015StochasticOW}, in deep learning settings \cite{Katharopoulos2018NotAS}, and in minibatches \cite{Csiba2018ImportanceSF}. As stated in these papers, importance sampling theoretically improves the convergence rate and is experimentally effective in reducing the training time and training loss. However, no generalization work has been done in a boosting setting.

\section{Algorithms for CNN as Weak Learner}
In this section, we provide a summary of BoostCNN and propose a new algorithm, subgrid CNN, which combines BoostCNN and the subgrid trick.
\subsection{Background: Standard BoostCNN}
We start with a brief overview of multiclass boosting. Given a sample $x_i\in\mathcal{X}$ and its class label $z_i\in\left\{1,2,\cdots,M\right\}$, multiclass boosting is a method that combines several multiclass predictors $g_t:\mathcal{X}\rightarrow \mathbb{R}^d$ to form a strong committee $f(x)$ of classifiers, i.e. $f(x)=\sum_{t=1}^N\alpha_t g_t(x)$ where $g_t$ and $\alpha_t$ are the weak learner and coefficient selected at the $t^{\mathrm{th}}$ boosting iteration. There are various approaches for multiclass boosting such as \cite{Hastie2009MulticlassA}, \cite{Mukherjee2013ATO}, \cite{Saberian2011MulticlassBT}; we use the GD-MCBoost method of \cite{Saberian2011MulticlassBT}, \cite{Moghimi2016BoostedCN} herein. For simplicity, in the rest of the paper, we assume that $d=M$.

Standard BoostCNN \cite{Moghimi2016BoostedCN} trains a boosted predictor $f(x)$ by minimizing the risk of classification
\begin{align}
\label{risk function}
    \mathcal{R}[f]=\mathrm{E}_{X,Z}\left[L(z,f(x)) \right]\approx \frac{1}{\left|\mathcal{D} \right|}\sum_{(x_i,z_i)\in\mathcal{D}}L(z_i,f(x_i)),
\end{align}
where $\mathcal{D}$ is the set of training samples and 
\begin{align*}
    L(z,f(x))=\sum_{j=1, j\neq z}^M e^{\frac{1}{2}\left[ \left\langle y_{z},f(x)\right\rangle -\left\langle y_j,f(x) \right\rangle \right]},
\end{align*}
given $y_k=\mathds{1}_k\in\mathbb{R}^M$, i.e. the $k^{\mathrm{th}}$ unit vector. The minimization is via gradient descent in a functional space. Standard BoostCNN starts with $f(x)=\mathbf{0}\in\mathbb{R}^d$ for every $x$ and iteratively computes the directional derivative of risk (\ref{risk function}), for updating $f(x)$ along the direction of $g(x)$
\begin{align}
    \delta \mathcal{R}[f;g]&=\left.\frac{\partial \mathcal{R}[f+\epsilon g]}{\partial \epsilon}\right|_{\epsilon=0}=-\frac{1}{2\left| \mathcal{D}\right|}\sum_{(x_i,z_i)\in\mathcal{D}}\sum_{j=1}^Mg_j(x_i)w_j(x_i,z_i)\nonumber\\
    &=-\frac{1}{2\left| \mathcal{D}\right|}\sum_{(x_i,z_i)\in\mathcal{D}} g(x_i)^Tw(x_i,z_i),
    \label{functional gradient}
\end{align}
where 
\begin{align}
\label{weight update}
    w_k(x,z)=\left\{\begin{matrix}
&-e^{-\frac{1}{2}\left [ f_{z}(x)-f_k(x) \right ]},\quad k\neq z\\ 
&\sum_{j=1,j\neq k}^M e^{-\frac{1}{2}\left [ f_{z}(x)-f_j(x) \right ]}, \quad k=z,
\end{matrix}\right.
\end{align}
and $g_j(x_i)$ computes the directional derivative along $\mathds{1}_j$. Then, standard BoostCNN selects a weak learner $g^*$ that minimizes (\ref{functional gradient}), which essentially measures the similarity between the boosting weights $w(x_i,z_i)$ and the function values $g(x_i)$. Therefore, the optimal network output $g^*(x_i)$ has to be proportional to the boosting weights, i.e. 
\begin{align}
\label{eq:grad-weight}
    g^*(x_i)=\beta w(x_i, z_i),
\end{align}
for some constant $\beta > 0$. Note that the exact value of $\beta$ is irrelevant since $g^*(x_i)$ is scaled when computing $\alpha^*$. Consequently, without loss of generality, we assume $\beta=1$ and convert the problem to finding a network $g(x)\in \mathbb{R}^M$ that minimizes the square error loss
\begin{align}
\label{weak learner train}
    \mathcal{L}(w,g)=\sum_{(x_i,z_i)\in\mathcal{D}}\left\|g(x_i)-w(x_i,z_i)\right\|^2.
\end{align}
After the weak learner is trained, BoostCNN applies a line search to compute the optimal step size along $g^*$,
\begin{align}
    \alpha^*=\argmin_{\alpha\in\mathbb{R}}\mathcal{R}[f+\alpha g^*].
    \label{boost parameter}
\end{align}
Finally, the boosted predictor $f(x)$ is updated as $f=f+\alpha^* g^*$.

\subsection{Subgrid BoostCNN}
When considering full-size images, BoostCNN using complex CNNs as weak learners is time-consuming and memory hungry. Consequently, we would like to reduce the size of the images to lower the running time and the memory requirement. A straightforward idea would be downsizing the images directly. A problem of this approach is that the noise would possibly spread out to later learners since a strong signal could be weakened during the downsize process. Another candidate for solving the aforementioned problem is randomly selecting pixels from the original images, however, the fluctuation of the performance of the algorithm would be significant especially when the images are sharp or have a lot of noise. In this paper, we apply the subgrid trick to each weak learner in BoostCNN. The remaining question is how to select a subgrid for each weak learner. Formally, a subgrid is defined by deleting a subset of rows and columns. Moreover, the processed images may not have the same size between iterations, which in turn requires that the new BoostCNN should allow each weak learner to have a at least different dimensions. However, that impedes reusing weak learner model parameters from one  weak learner iterate to next.

In order to address these issues, we first separate a standard deep CNN into two parts. We call all  layers such as convolutional layers and pooling layers, except the last fully-connected (FC) layers, the {\it feature extractor}. In contrast, we call the last FC layers the {\it classifier}. Furthermore, we refer to $g_0$ as the basic weak learner and all the succeeding $g_t$ as the additive weak learners. Subgrid BoostCNN defines an importance index for each pixel $(j,k)$ in the image as 
\begin{align}
\label{importance}
    I_{j,k}=\frac{1}{\left|\mathcal{D}\right|}\sum_{(x_i,z_i)\in\mathcal{D}}\sum_{c\in C}\left|\frac{\partial \mathcal{L}(w,g)}{\partial x_i^{j,k,c}}\right|,
\end{align}
where $x_i^{j,k,c}$ denotes pixel $(j,k)$ in channel $c$ from sample $i$ and $C$ represents the set of all channels. The importance index of a row, column is a summation of the importance indexes in the row, column divided by the number of columns, rows, respectively. This importance index is computed based on the residual of the current predictor. Therefore, a larger importance value means a larger adjustment is needed for this pixel at the current iterate. The algorithm uses the importance index generated based on the feature extractor of the incumbent weak learner and the classifier from $g_0$ to conduct subgrid selection. The selection strategy we apply in the algorithm is deleting less important columns and rows, which eventually provides the important subgrid. After the subgrid is selected, subgrid BoostCNN creates a new tensor $x_i^t$ at iterate $t$, and then feeds it into an appropriate feature extractor followed by a proper classifier. The modified minimization problem becomes
\begin{align}
\label{subgrid weak learner train}
  \mathcal{L}(w,g)=\sum_{(x_i,z_i)\in\mathcal{D}}\left\|g(x_i^t)-w(x_i,z_i)\right\|^2, 
\end{align}
where the modified boosting classifier is 
\begin{align}
    f(x)=\sum_{t=1}^N\alpha_tg_t(x^t).
    \label{subgrid classifier}
\end{align}
In this way, subgrid BoostCNN dynamically selects important subgrids based on the updated residuals. Moreover, subgrid BoostCNN is able to deal with inputs of different sizes by applying different classifiers. Furthermore, we are allowed to pass the feature extractor's parameters from the previous weak learner since the feature extractor is not restricted to the input size. The proposed algorithm (subgrid BoostCNN) is summarized in Algorithm \ref{alg:subBoostCNN}.

\begin{algorithm}[H]
  \caption{subgrid BoostCNN}
  \label{alg:subBoostCNN}
  \begin{algorithmic}[1]
  \Inputs{number of classes $M$, number of boosting iterations $N_b$, shrinkage parameter $\nu$, dataset $\mathcal{D}=\left\{(x_1,z_1),\cdots,(x_n,z_n)\right\}$ where $z_i\in\left\{1,\cdots,M\right\}$ is the label of sample $x_i$, and $0<\sigma<1$}
    \Initialize{set $f(x)=\mathbf{0}\in\mathbb{R}^M$, $P_0=\left\{(j,k)\vert (j,k)\mathrm{\,is\,a\,pixel\,in\,}x_i \right\}$ }
    \State compute $w(x_i,z_i)$ for all $(x_i,z_i)$, using (\ref{weight update}) \label{basic start}
    \State train a deep CNN $g_0^*$ to optimize (\ref{weak learner train})\label{basic end}
     \State $f(x) = g_0^*$
    \For{t = $1,2,\cdots$, $N_b$ }
    \State update importance index $I_{j,k}$ for $(j,k)\in P_{t-1}$, using (\ref{importance})\label{generate matrix}
        \State select the subgrid based on $\sigma$ fraction of rows and columns with highest importance index and let $P_t$ be the set of selected pixels; form a new tensor $x_i^t$ for each sample $i$\label{select subgrid}
        \State construct a new proper weak learner architecture\label{construct wl}
        \State compute $w(x_i,z_i)$ for all $i$, using (\ref{weight update}) and (\ref{subgrid classifier})\label{subgrid start}
    \State train a deep CNN $g_t^*$ to optimize (\ref{subgrid weak learner train})
    \State find the optimal coefficient $\alpha_t$, using (\ref{boost parameter}) and (\ref{subgrid classifier})\label{subgrid end}
     \State $f(x) = f(x) + \nu\alpha_t g_t^*$\label{subgrid update}
      \EndFor
      \State \textbf{end for}
  \end{algorithmic}
\end{algorithm}
Subgrid BoostCNN starts by initializing $f(x)=\mathbf{0}\in\mathbb{R}^M$. The algorithm first generates a full-size deep CNN as the basic weak learner, which uses the full image in steps~\ref{basic start}-\ref{basic end}. After the basic weak learner $g_0^*$ is generated, in each iteration, subgrid BoostCNN first updates the importance index $I_{j,k}$ for each pixel $(j,k)$, which has been used in the preceding iterate at step~\ref{generate matrix}. In order to mimic the loss of the full-size image, although we only update the importance indexes for the pixels which have been used in the last iterate, we feed the full-size tensor to the deep CNN $g$ to compute the importance index. The deep CNN $g$ used in (\ref{importance}) to compute the importance value is constructed by copying the feature extractor from the preceding weak learner followed by the classifier in the basic weak learner $g_0^*$. Next, by deleting less important rows and columns based on $I_{j,k}$, which contain $1-\sigma$ fraction of pixels, it finds the most important subgrid having $\sigma$ fraction of pixels at position $P_t$ based on the importance index $I_{j,k}$, and forms a new tensor $x_i^t$ in step~\ref{select subgrid}. Note that $P_t$ is not necessary to be a subset of $P_{t-1}$ and actually is rarely to be a subset of $P_{t-1}$. This only happens when the highest importance index at iterate $t$ is also the highest score at iterate $t-1$. Next, a new additive weak learner is initialized by borrowing the feature extractor from the preceding weak learner $g^*_{t-1}$ followed by a randomly initialized FC layer with the proper size in step~\ref{construct wl}. Once the additive weak learner is initialized, subgrid BoostCNN computes the boosting weights, $w(x)\in\mathrm{R}^M$ according to (\ref{weight update}) and (\ref{subgrid classifier}), trains a network $g_t^*$ to minimize the squared error between the network output and boosting weights using (\ref{subgrid weak learner train}), and finds the boosting coefficient $\alpha_t$ by minimizing the boosting loss (\ref{boost parameter}) in steps~\ref{subgrid start}-\ref{subgrid end}. Lastly, the algorithm adds the network to the ensemble according to $f(x) = f(x) + \nu\alpha_t g_t^*$ for $\nu\in[0,1]$ in step~\ref{subgrid update}.

\section{Algorithms for Transformer as Weak Learner}
In this section, we propose three algorithms combining boosting and transformers from different perspectives. We assume a BERT-like bidirectional transformer classifier \cite{Devlin2019BERTPO} \cite{liu2019roberta}. The first token of each sequence is a special classification token, and the corresponding final hidden state output of this token is used as the aggregated representation for the classification.

\subsection{Standard BoostTransformer}
Inspired by BoostCNN, we propose BoostTransformer which combines boosting and transformers (encoder) together. For a sequence classification problem, we are given a sample $x_i\in\mathcal{X}$, which contains a sequence of tokens, and its class label $z_i\in\left\{1,2,\cdots,M\right\}$. The risk function, the functional gradient and the optimal boosting coefficient $\alpha^t$ are exactly the same as those in (\ref{risk function}), (\ref{functional gradient}), and (\ref{boost parameter}), respectively. The algorithm follows standard gradient boosting machine.

\subsection{Subsequence BoostTransformer}
Combining the subgrid trick and BoostTransformer means applying the subgrid trick to each weak learner in BoostTransformer. Different from deep CNNs, transformers are able to deal with sequences of any length, thus, there is no issue when transferring information from the current weak learner to the succeeding weak learner. Similar to subgrid BoostCNN, we denote $g_0$ as the basic weak learner, which deals with the whole dataset, and all the succeeding $g_t$'s as the additive weak learners. Moreover, subsequence BoostTransformer defines an importance index for each token $\mathcal{w}$ in the vocabulary based on the attention distribution. More precisely, the importance value of token $\mathcal{w}$ is computed by adding two parts; the first part is the importance of the token $\mathcal{w}$ itself, and the second part is the importance of token $\mathcal{w}$ to the remaining tokens in the same sample. In an $L$-layer transformer for a sequence $x$ of length $s$ (following \cite{Devlin2019BERTPO} we assume that the first token in $x$ is a placeholder, which indicates that the corresponding token in the final layer is used as the embedding for classification), and positions $1\leq i,j\leq s$, and layer $k$ for $1\leq k\leq L$, let the attention from position $i$ to position $j$ between layer $k-1$ and $k$ be denoted by $a(i,j;k;x)$. We have $\sum_{j=1}^s a(j,i;k;x)=1$ for every $i,k,x$. Then, given a transformer with $L$ layers, the self-importance of token $\mathcal{w}$ in position $p$ in a sample $x_i$ is
\begin{align}
\label{importance_self}
I^S(\mathcal{w},x_i)=\left(\prod_{k=1}^{L-1} a(p,p;k;x_i)\right)\cdot a(p,1;L;x_i)\approx a(p,1;L;x_i),
\end{align}
The importance of token $\mathcal{w}$ to others is 
\begin{align}
\label{importance_rest}
I^R(\mathcal{w},x_i)=\left[\prod_{k=1}^{L-1}\max_{j,j\neq p} a(p_{k-1},j;k;x_i)\right]\cdot a(p_{L-1},1;L;x_i),
\end{align}
where $p_{k-1}=\mathrm{argmax}_{j,j\neq p}a(p_{k-2},j;k-1;x_i)$  for $k=2,3,\cdots, L-1$, and $p_0=p$. The first term computes the product of the maximum attention values through the path which does not contain $p$ until the second to last layer. For the second term, as it has been shown in \cite{Devlin2019BERTPO}, the classification layer only takes the $1$st position of the last transformer layer which is corresponding to the classification token, therefore, the formula in (\ref{importance_rest}) does not check all possible attention distributions; instead, it counts the attention value from the position $p_{L-1}$ to the $1$st position in the last transformer layer directly. After the aforementioned importance values are computed, the importance value of the vocabulary word $\hat{\mathcal{w}}$ is
\begin{align}
\label{importance_agg}
    I(\hat{\mathcal{w}})=\sum_{
    \begin{matrix}
x_i,\mathcal{w}\in x_i \\
\mathcal{w} = \hat{\mathcal{w}}
\end{matrix}} \left(I^S(\mathcal{w},x_i)+ I^R(\mathcal{w},x_i)\right).
\end{align}
Then, the algorithm uses the importance index to select the most important tokens. After the tokens are selected, subsequence BoostTransformer creates a new sample $x_i^t$ at iterate $t$, which contains only the important tokens, and is used by the weak learner. The modified minimization problem and the boosting weak learner are explicitly presented in (\ref{subgrid weak learner train}) and (\ref{subgrid classifier}), respectively. The proposed algorithm (subsequence BoostTransformer) is summarized in Algorithm \ref{alg:subBoostTrans}.

Different from standard BoostTransformer, subsequence BoostTransformer first reviews the whole dataset in steps~\ref{subT:basic start}-\ref{subT:basic end} and generates the basic weak learner $g_0^*$. Once the basic weak learner is created, in each iteration, subsequence BoostTransformer first updates the attention-based importance vector $I_{\mathcal{w}}$ for any $\mathcal{w}\in V_{t-1}$ in step~\ref{subT:generate matrix}, and selects $\sigma$ fraction of the tokens to form the vocabulary set $V_t$, and lastly constructs a new sample $x_i^t$ by deleting any tokens not in $V_t$ in step~\ref{subT:select subgrid}. After the new sample $x_i^t$ is constructed, subsequence BoostTransformer initializes the weights of the current transformer by using the weights in $g_{t-1}^*$ and trains the transformer with $x_i^t$ to minimize the squared error in (\ref{subgrid weak learner train}) in steps~\ref{subT:subgrid start}-\ref{subT:subgrid end}. Lastly, the algorithm finds the boosting coefficient $\alpha_t$ by minimizing (\ref{boost parameter}) in step~\ref{subT:coeff} and adds the additive weak learner to the ensemble in step~\ref{subT:subgrid update}.
\begin{algorithm}[H]
  \caption{subsequence BoostTransformer}
  \label{alg:subBoostTrans}
  \begin{algorithmic}[1]
  \Inputs{number of classes $M$, number of boosting iterations $N_b$, shrinkage parameter $\nu$, dataset $\mathcal{D}=\left\{(x_1,z_1),\cdots,(x_n,z_n)\right\}$ where $z_i\in\left\{1,\cdots,M\right\}$ is the label of sample $x_i$, and $0<\sigma<1$}
    \Initialize{set $f(x)=\mathbf{0}\in\mathbb{R}^M$, $V_0=\left\{\mathcal{w}\vert \mathcal{w}\in x_i  \mathrm{\,\,for\,\, some\,\,}x_i\right\}$ }
    \State compute $w(x_i,z_i)$ for all $(x_i,z_i)$, using (\ref{weight update}) \label{subT:basic start}
    \State train a transformer $g_0^*$ to optimize (\ref{weak learner train})\label{subT:basic end}
     \State $f(x) = g_0^*$
    \For{t = $1,2,\cdots ,$ $N_b$ }
    \State update importance values $I_{\mathcal{w}}$ for $\mathcal{w}\in V_{t-1}$, using (\ref{importance_self}), (\ref{importance_rest}) and (\ref{importance_agg})\label{subT:generate matrix}
    \State form $V_t\subset V_{t-1}$ with $\frac{|V_t|}{|V_{t-1}|}\approx\sigma$ and $I(\mathcal{w})>I(\mathcal{w}') \forall \mathcal{w}\in V_t,\mathcal{w}' \in V_{t-1}\setminus V_t$   and form a new sample $x_i^t$ for each sample $i$\label{subT:select subgrid}
    \State compute $w(x_i,z_i)$ for all $i$, using (\ref{weight update}) and (\ref{subgrid classifier})\label{subT:subgrid start}
    \State train a transformer $g_t^*$ to optimize (\ref{subgrid weak learner train})\label{subT:subgrid end}
    \State find the optimal coefficient $\alpha_t$, using (\ref{boost parameter}) and (\ref{subgrid classifier})\label{subT:coeff}
     \State $f(x) = f(x) + \nu\alpha_t g_t^*$\label{subT:subgrid update}
      \EndFor
      \State \textbf{end for}
  \end{algorithmic}
\end{algorithm}
\subsection{Importance-sampling-based BoostTransformer}
Importance sampling, a strategy for preferential sampling of more important samples capable of accelerating the training process, has been well studied in stochastic gradient descent (SGD) \cite{Alain2015VarianceRI}. However, there is virtually no existing work combining the power of importance sampling with the strength of boosting. Motivated by the phenomenon that overfitting appears early in standard BoostTransformer, we propose importance-sampling-based BoostTransformer, which combines importance sampling and BoostTransformer. Importance-sampling-based BoostTransformer mimics importance sampling SGD by introducing a new loss function and computing a probability distribution for drawing samples. Similarly, importance-sampling-based BoostTransformer computes a probability distribution in each iteration, and draws a subset of samples to train the weak learner based on the distribution. The probability distribution is
\begin{align}
\label{importance sample distribution}
    P(I=i)=\frac{\left\|w(x_i,z_i)\right\|}{\sum_{(x_j,z_j)\in\mathcal{D}}\left\|w(x_j,z_j)\right\|},
\end{align}
which yields the new loss function for a subset of samples ${\mathcal{I}}$ to be
\begin{align}
\label{importance sample train loss}
    \bar{\mathcal{L}}_{\mathcal{I}}(w,g)=\sum_{(x_i,z_i)\in\mathcal{I}}\frac{1}{\left|\mathcal{D}\right|P(I=i)}\left\|g(x_i)-w(x_i,z_i)\right\|^2.
\end{align}
To any minimization algorithm one would typically use.We then apply any optimization algorithm with respect to (\ref{importance sample train loss}) (by further using mini-batches or importance sampling).

The entire algorithm is exhibited in Algorithm \ref{alg:impBoostTrans}.
\begin{algorithm}[H]
  \caption{importance-sampling-based BoostTransformer}
  \label{alg:impBoostTrans}
  \begin{algorithmic}[1]
  \Inputs{number of classes $M$, number of boosting iterations $N_b$, shrinkage parameter $\nu$, dataset $\mathcal{D}=\left\{(x_1,z_1),\cdots,(x_n,z_n)\right\}$ where $z_i\in\left\{1,\cdots,M\right\}$ is the label of sample $x_i$, and $0<\sigma<1$}
    \Initialize{set $f(x)=\mathbf{0}\in\mathbb{R}^M$}
    \State compute $w(x_i,z_i)$ for all $x_i$, using (\ref{weight update}) \label{imT:basic start}
    \State train a transformer $g_0^*$ to optimize (\ref{weak learner train})\label{imT:basic end}
     \State $f(x) = g_0^*$
    \For{t = $1,2,\cdots,$ $N_b$ }
    \State compute probability distribution $P_t$, using (\ref{importance sample distribution})\label{imT:importance sample}
        \State draw independently  $|\mathcal{I}^t|$ samples, which is $\sigma$ fraction of the samples, based on $P_t$\label{imT:select sample}
        \State compute $w(x_i,z_i)$ for $(x_i,z_i)\in \mathcal{I}^t$, using (\ref{weight update}) \label{imT:subgrid start}
    \State train a transformer $g_t^*$ to optimize (\ref{importance sample train loss}) on $\mathcal{I}^t$\label{imT:subgrid end}
    \State find the optimal coefficient $\alpha_t$, using (\ref{boost parameter}) on $\mathcal{I}^t$\label{imT:coeff}
     \State $f(x) = f(x) + \nu\alpha_t g_t^*$\label{imT:subgrid update}
      \EndFor
      \State \textbf{end for}
  \end{algorithmic}
\end{algorithm}
Importance-sampling-based BoostTransformer starts with learning the full-size dataset and training a basic weak learner in steps~\ref{imT:basic start}-\ref{imT:basic end}. In each iteration, the algorithm first computes the probability distribution $P_t$ in step~\ref{imT:importance sample} and selects a subset $\mathcal{I}^t$ of samples based on the distribution in step~\ref{imT:select sample}. Once the dataset is created, it computes the weights and trains a transformer by using the unbiased loss function (\ref{importance sample train loss}), following by finding an optimal boosting coefficient in steps~\ref{imT:subgrid start}-\ref{imT:subgrid update}.

In the rest of this section, we provide all analysis of the optimal probability distribution in importance-sampling-based BoostTransformer. Given current aggregated classifier $f_{t-1}=\sum_{i=1}^{t-1}g_{i}^*$, let us define the expected training progress attributable to iteration $t$ as
\begin{align*}
    \mathbb{E}_{P_{t}}\left[ \Delta^{(t)} \right]=\left\|f_{t-1}-f^*\right\|^2-\mathbb{E}_{P_{t}}\left[\left.\left\| f_{t}-f^*\right\|^2 \right|\mathcal{F}^{t-1}\right].
\end{align*}
Here $f^*$ denotes the solution to (\ref{risk function}), and the expectation is taken over the probability distribution $P_t$, and $\mathcal{F}^{t-1}$ contains the whole history of the algorithm up until iterate $t-1$. We assume that gradient sampling is unbiased. Inspired by the work in \cite{Zhao2015StochasticOW}, we prove that the optimal probability distribution is proportional to the boosting weight at each iteration.
\begin{theorem}\label{thm:1}
In $\max_{P_{t}}$ $\mathbb{E}_{P_{t}}\left[ \Delta^{(t)}\right]$, the optimal distribution for importance-sampling-based BoostTransformer to select each sample $i$ is proportional to its “boosting weight norm:” , i.e. (\ref{importance sample distribution}).
\end{theorem}
\begin{proof}
See Appendix \hyperref[pf:thm1]{A}.
\end{proof}
Based on the fact that the new loss function with respect to the probability distribution is unbiased, we discover that maximizing the improvement of the boosting algorithm is equivalent to minimizing the functional gradient variance. By applying Jensen's inequality, the optimal probability distribution is essentially proportional to the boosting weights, which are easy to obtain, in boosting algorithms.  
\section{Experimental Study}
In this section, we first compare subgrid BoostCNN with standard BoostCNN and deep CNNs, next, we compare the standard transformer, BoostTransformer, subsequence BoostTransformer and importance-sampling-based BoostTransformer in the second half of the section. We conduct experiments on three different datasets for both CNN related and transformer related algorithms. From all of these datasets, we study the performance of the boosting technique, the subgrid trick and the importance sampling strategy. All the algorithms are implemented in Python with PyTorch \cite{paszke2017automatic}. Training is conducted on an NVIDIA Titan XP GPU.
\subsection{Image}
In this subsection, we illustrate properties of the proposed subgrid BoostCNN and compare its performance with other methods on several image classification tasks. In subgrid BoostCNN, the risk function (\ref{risk function}) we employ is cross entropy, and the input of each weak learner is an image with 3 channels which can be handled by standard Conv2d functions in PyTorch. Meanwhile, we implement the subgrid strategy based on (\ref{importance}) with respect to each pixel $(j,k)$. We delete approximately $10\%$ of the rows and columns, which implies $\sigma = 81\%$ on the total number of pixels, and fix the shrinkage parameter $\nu$ to be $0.02$. In each weak learner, we apply the ADAM algorithm with the learning rate of $0.0001$ and weight decay being $0.0001$.

We consider CIFAR-10 \cite{Krizhevsky2009LearningML}, SVHN \cite{Netzer2011ReadingDI} and ImageNetSub \cite{Deng2009ImageNetAL} datasets as shown in Table \ref{tb:data1}. For the last dataset, since the original ImageNet dataset is large and takes significant amount of time to train, we select a subset of samples from the original ImageNet dataset. More precisely, we randomly pick 100 labels and select the corresponding samples from ImageNet, which consists of $124,000$ images for training and $10,000$ images for testing. We denote it as ImageNetSub. Data preprocessing consists of three steps: 1. random resizing and cropping with output size $224\times 224$, scale uniformly sampled from [0.08, 1.0] and make the aspect ratio uniformly sampled from [0.75, 1.33]; 2. random horizontal flipping with  flipping probability $0.5$; 3. normalization for each channel.
\begin{table}[H]
\centering
\begin{tabular}{|l|r|r|}
\hline
{\color[HTML]{333333} }         & \multicolumn{1}{c|}{{\color[HTML]{333333} Number of Training/Testing Samples}} & \multicolumn{1}{l|}{Number of Classes} \\ \hline
{\color[HTML]{333333} CIFAR-10} & \cellcolor[HTML]{FFFFFF}{\color[HTML]{333333} 50k/10k}            & 10                                     \\ \hline
{\color[HTML]{333333} SVHN}     & \cellcolor[HTML]{FFFFFF}{\color[HTML]{333333} 73k/26k}           & 10                                     \\ \hline
{\color[HTML]{333333} ImageNetSub} & \cellcolor[HTML]{FFFFFF}{\color[HTML]{333333} 124k/10k}           & 100                                    \\ \hline
\end{tabular}
\caption{Image Datasets}\label{tb:data1}
\end{table}

For training, we employ three different deep CNNs, which are ResNet-18, ResNet-50 and ResNet-101. For each combination of dataset/CNN, we first train the deep CNN for a certain number of epochs, and then initialize the weights in the basic weak learner for the boosting algorithms as the weights in the deep CNN. In the subgrid BoostCNN experiments, we use $10$ CNN weak learners. We train each weak learner for 15 epochs. For comparison, we train BoostCNN, the ensemble method (without boosting weight update and always using all features) denoted by e-CNN and the subgrid ensemble method named as subgrid e-CNN (without boosting weight update in step~\ref{subgrid start} in Algorithm \ref{alg:subBoostCNN}) for 10 iterates as well. Notice that subgrid e-CNN essentially mimics random forests. We also train the single deep CNN for 150 epochs to represent approximately the same computational effort as training 10 CNN weak learners for 15 epochs. 

\begin{figure}[H]
\centering
\begin{minipage}{.5\textwidth}
  \centering
  \includegraphics[width=.95\linewidth]{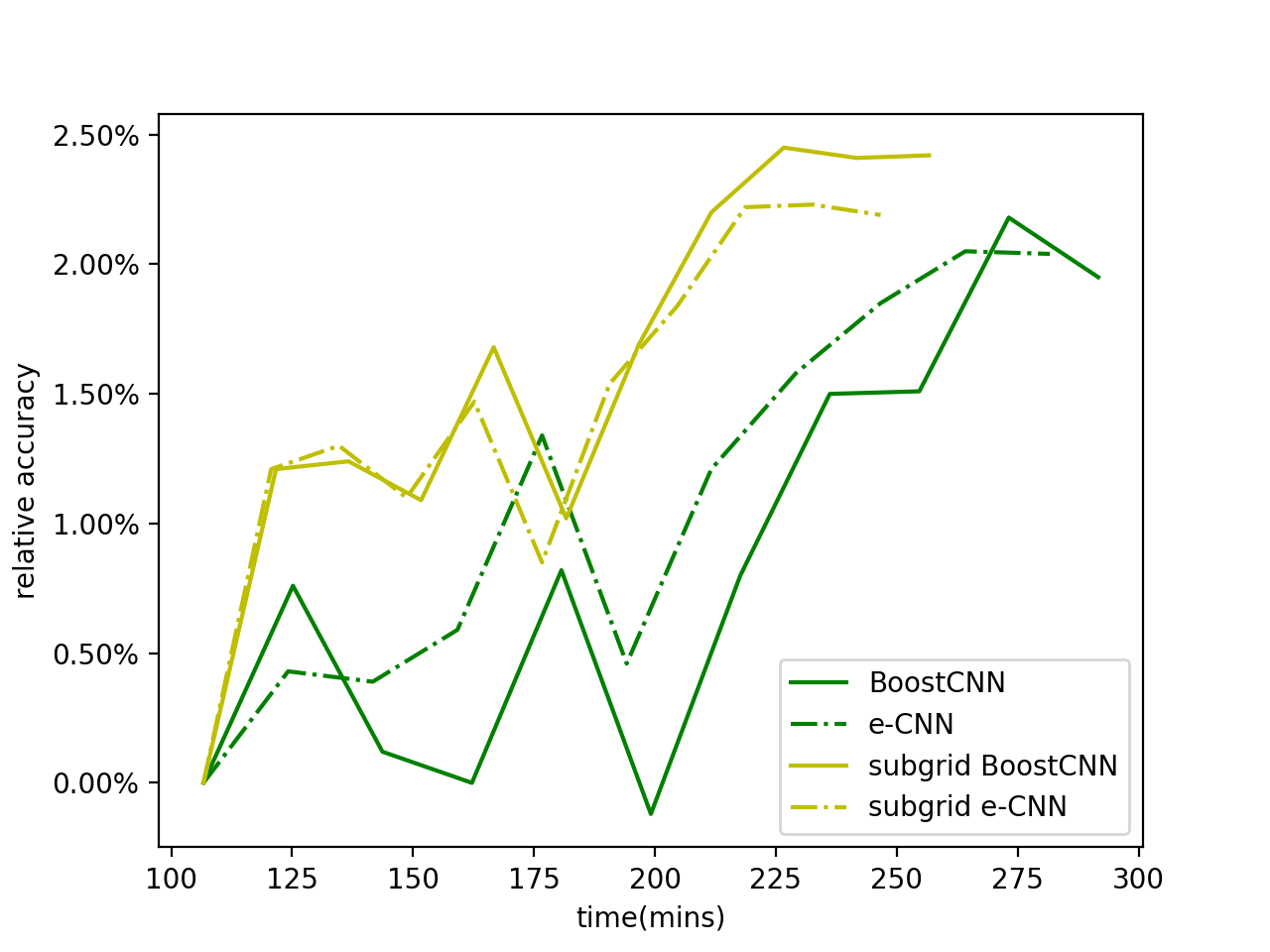}
  \captionof{figure}{ResNet-18 on CIFAR-10}
  \label{fig:18-cifar}
\end{minipage}%
\begin{minipage}{.5\textwidth}
  \centering
  \includegraphics[width=.95\linewidth]{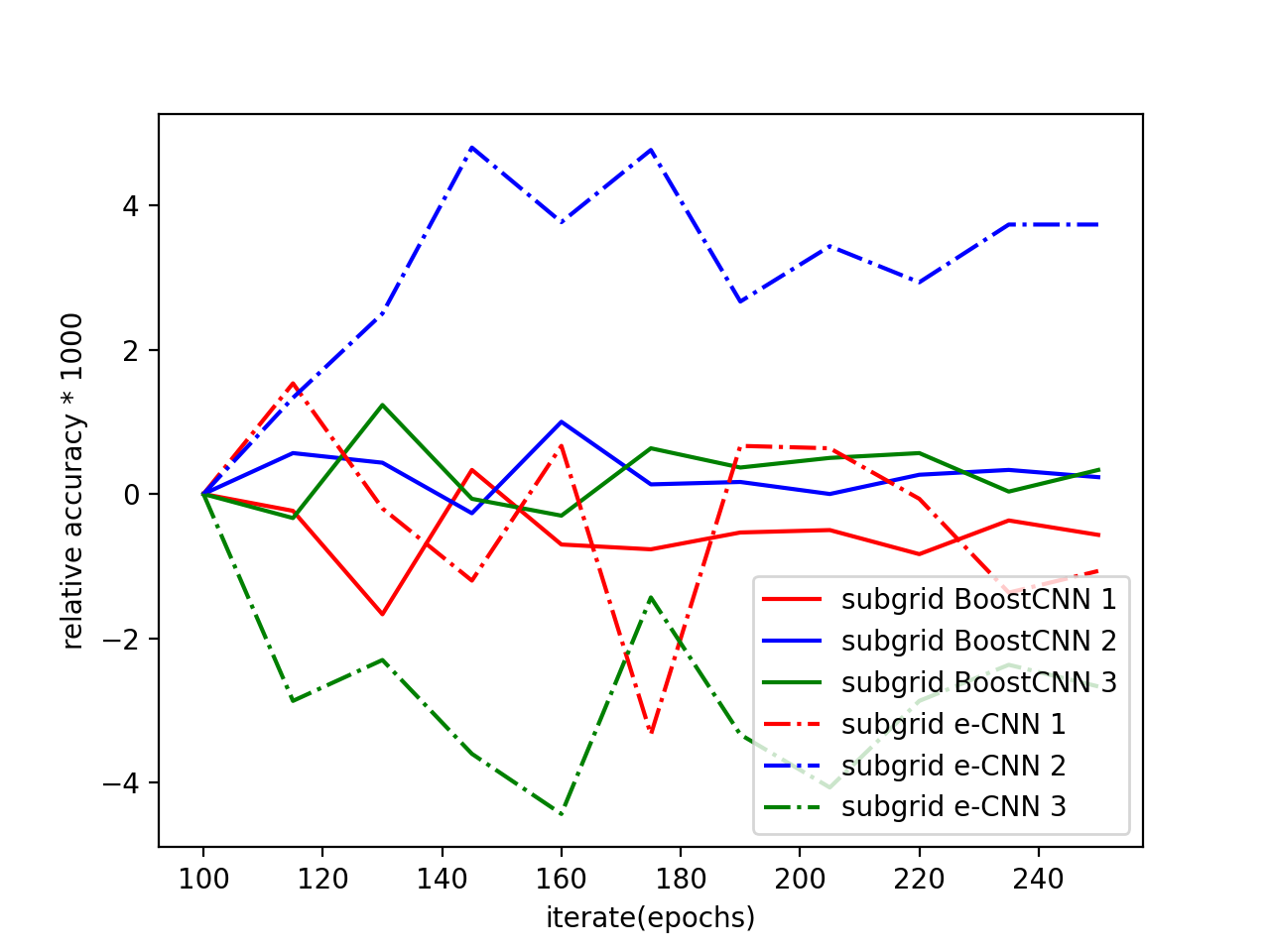}
  \captionof{figure}{Different Seeds}
  \label{fig:18-cifar-seed}
\end{minipage}
\begin{minipage}{.5\textwidth}
  \centering
  \includegraphics[width=.95\linewidth]{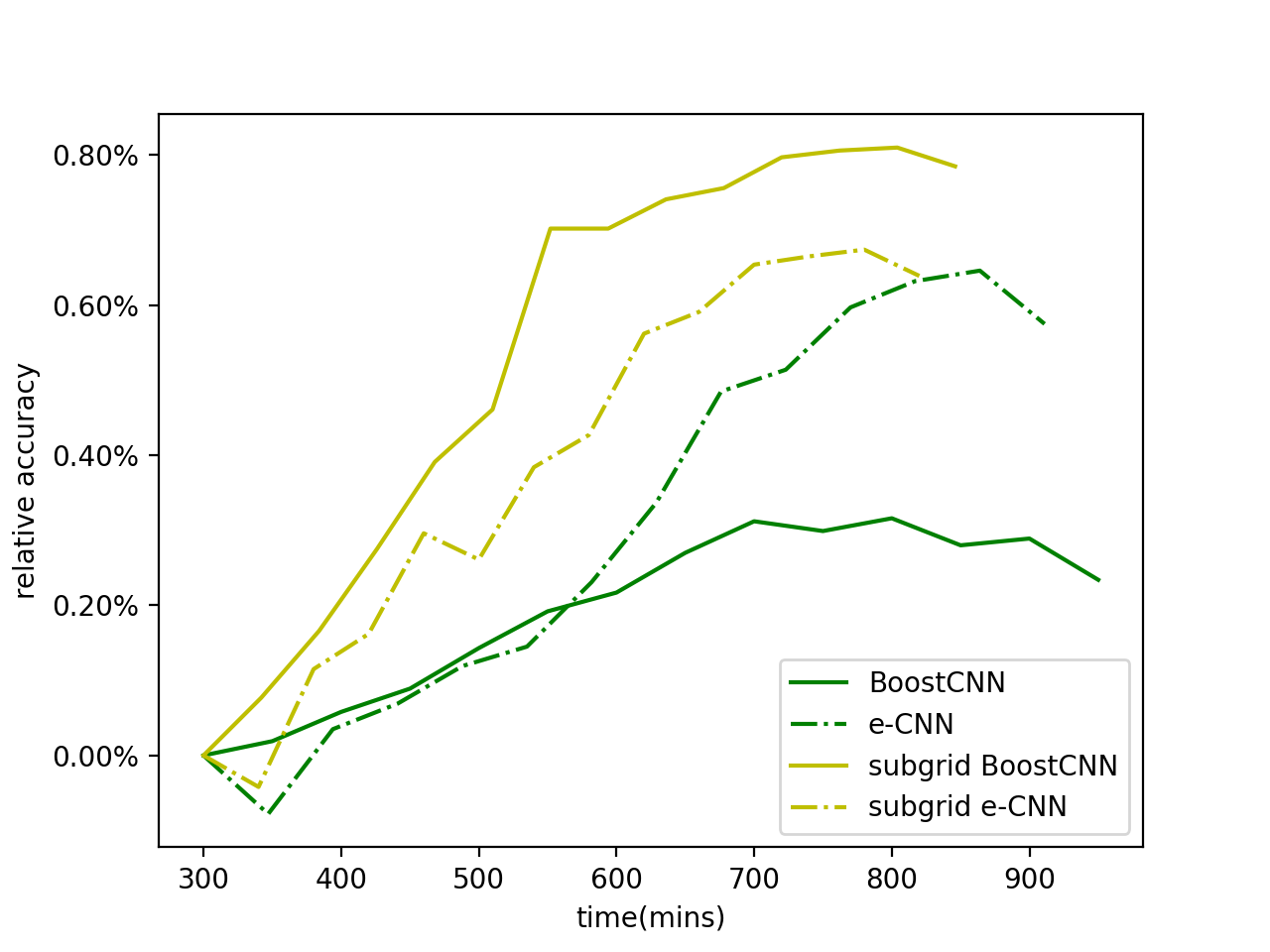}
  \captionof{figure}{ResNet-18 on SVHN}
  \label{fig:18-svhn}
\end{minipage}%
\begin{minipage}{.5\textwidth}
  \centering
  \includegraphics[width=.95\linewidth]{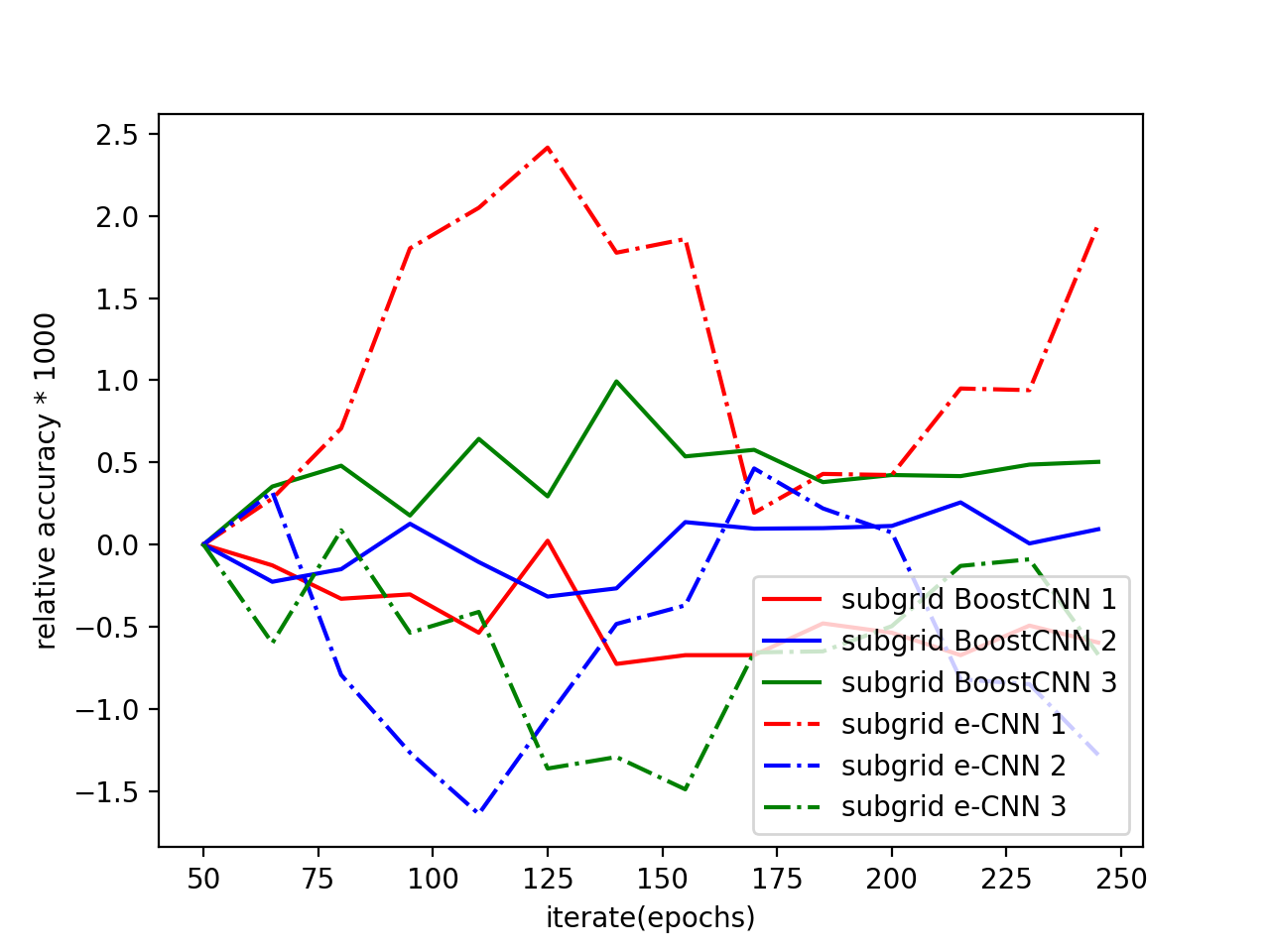}
  \captionof{figure}{Different Seeds}
  \label{fig:18-svhn-seed}
\end{minipage}
\begin{minipage}[t]{.5\textwidth}
  \centering
  \includegraphics[width=.95\linewidth]{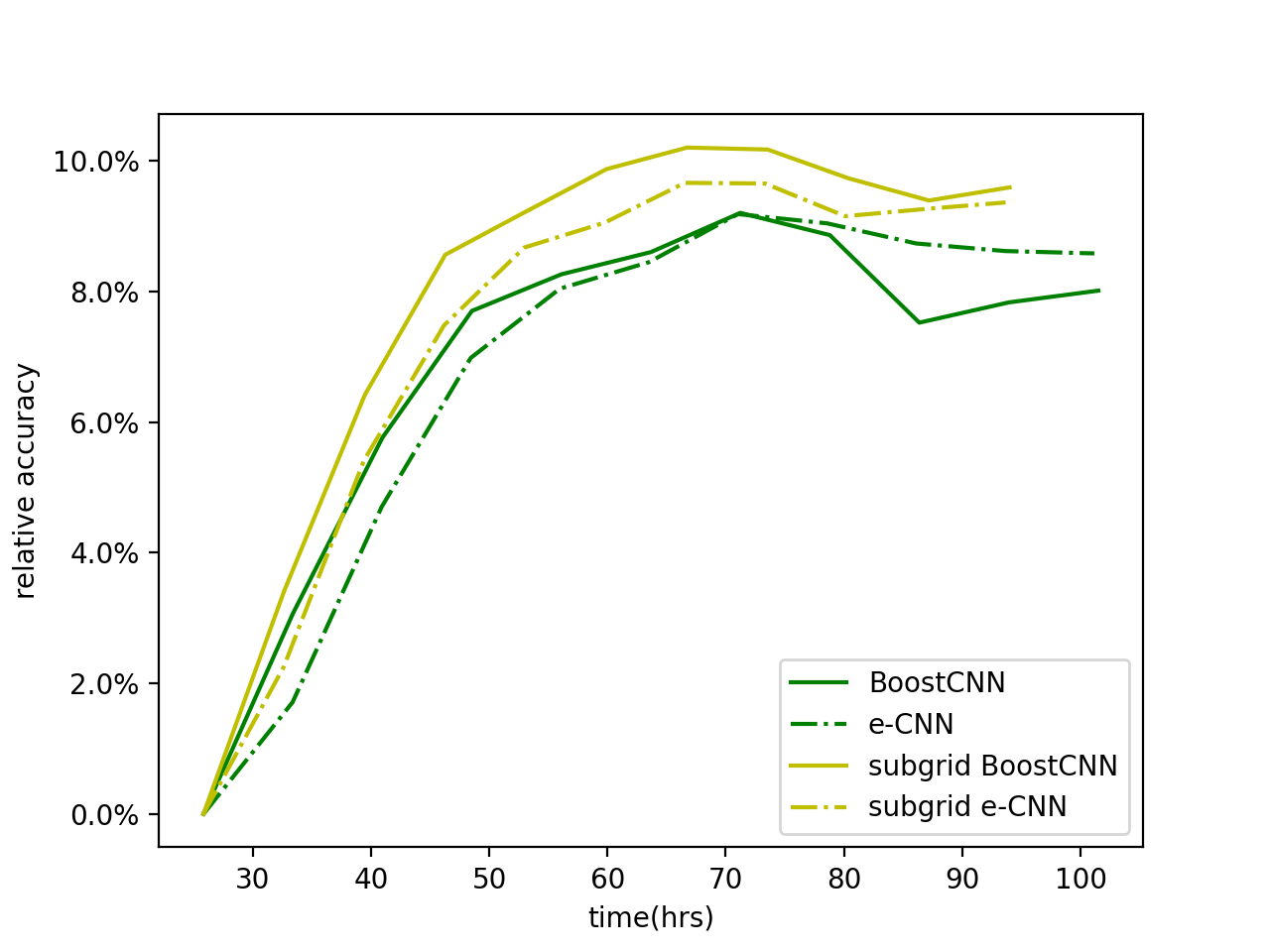}
  \captionof{figure}{ResNet-18 on ImageNetSub}
  \label{fig:18-image}
\end{minipage}%
\begin{minipage}[t]{.5\textwidth}
  \centering
  \includegraphics[width=.95\linewidth]{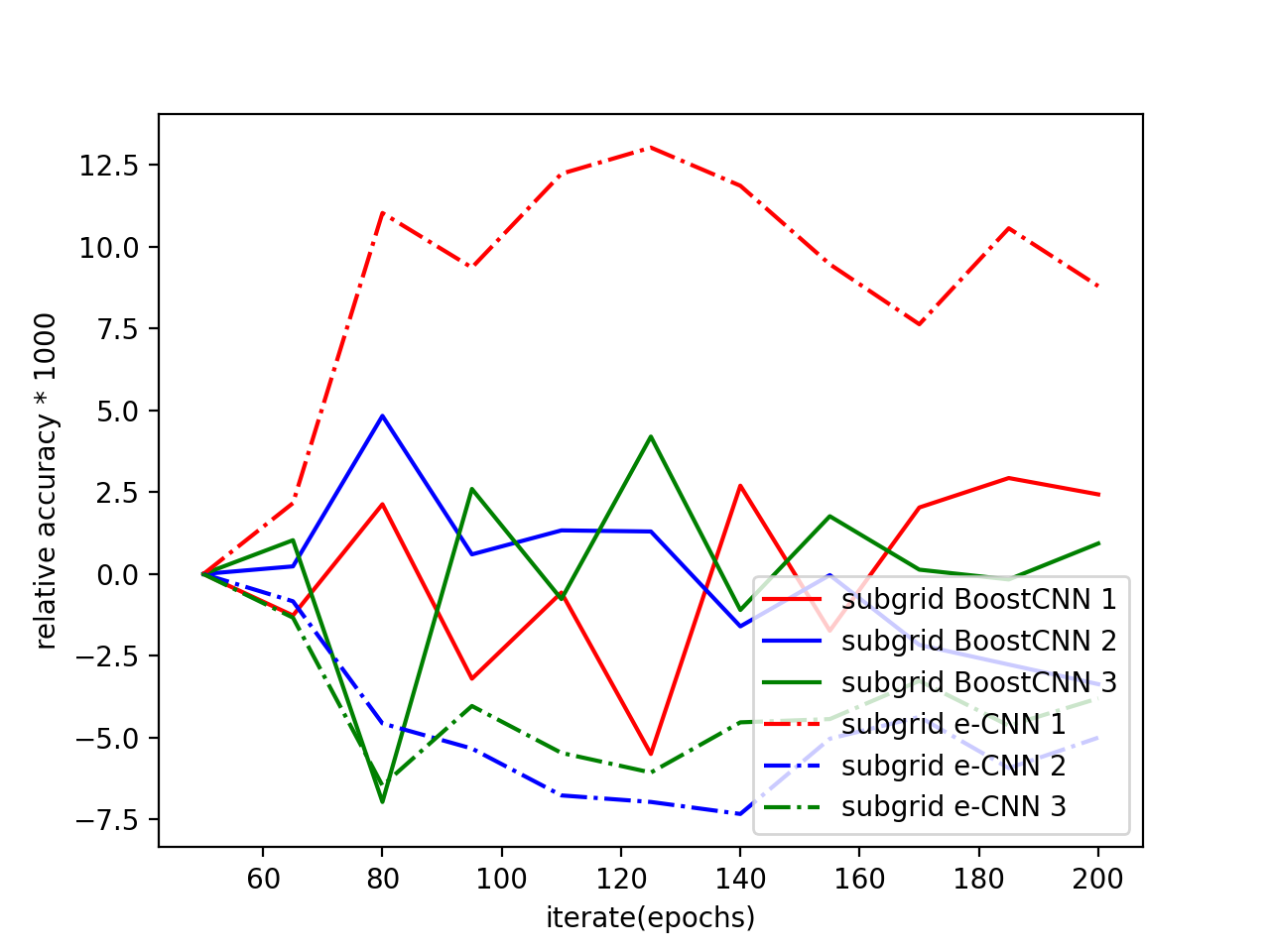}
  \captionof{figure}{Different Seeds}
  \label{fig:18-image-seed}
\end{minipage}
\end{figure}

\begin{figure}[H]
\centering
\begin{minipage}{.51\textwidth}
  \centering
  \includegraphics[width=.95\linewidth]{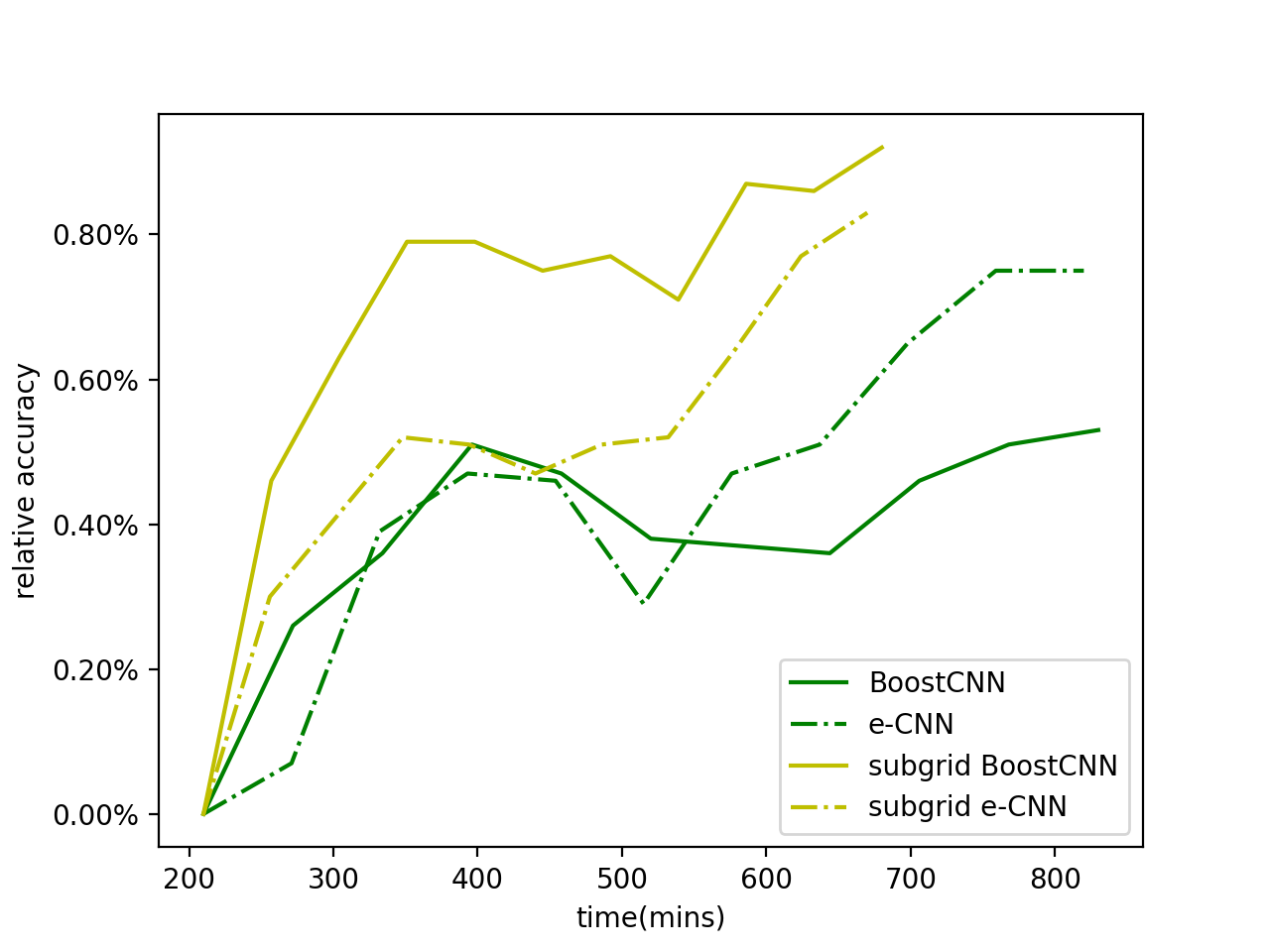}
  \captionof{figure}{ResNet-50 on CIRFAR-10}
  \label{fig:50-cifar}
\end{minipage}%
\begin{minipage}{.49\textwidth}
  \centering
  \includegraphics[width=1\linewidth]{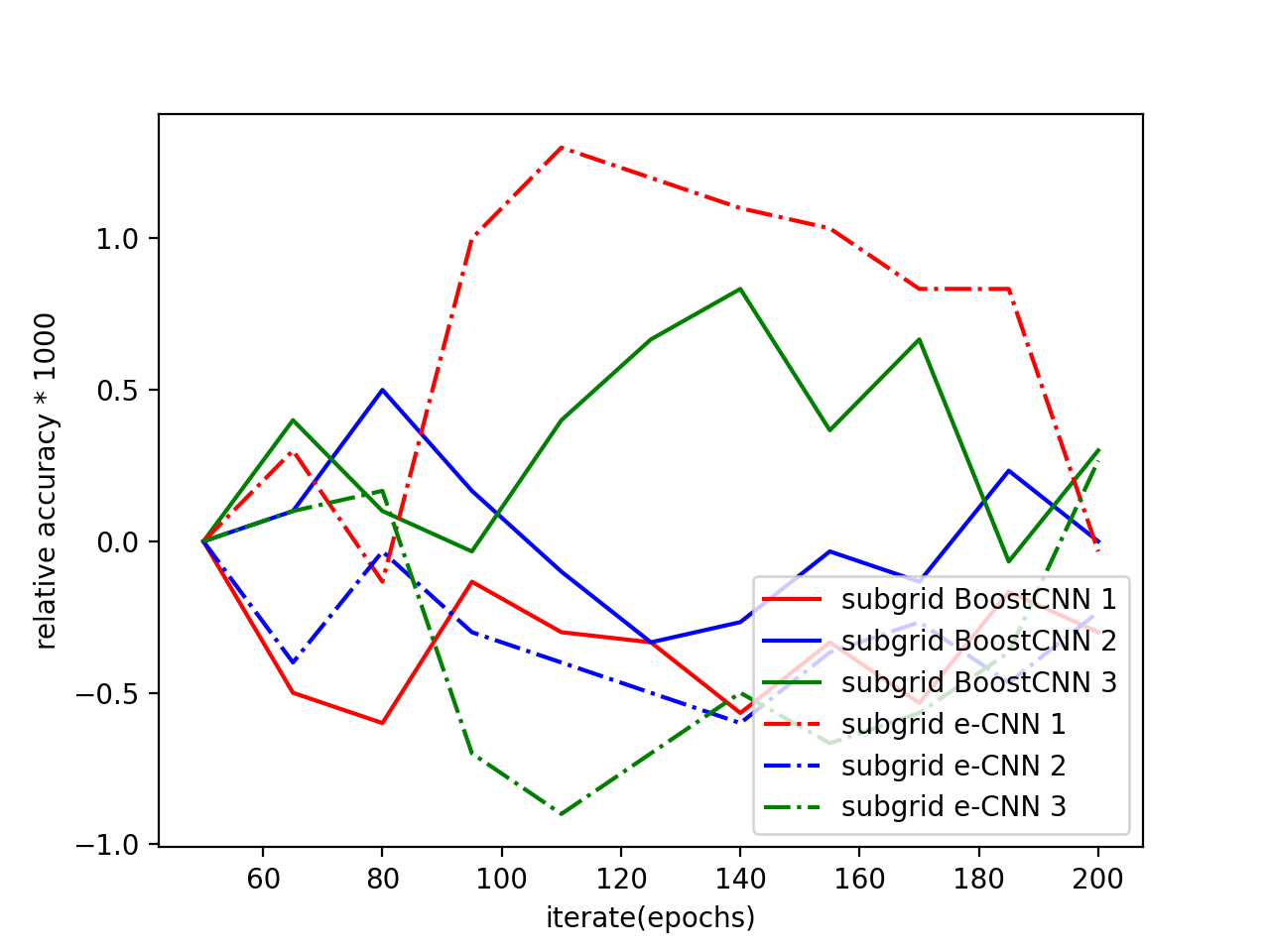}
  \captionof{figure}{Different Seeds}
  \label{fig:50-cifar-seed}
\end{minipage}
\begin{minipage}[t]{.5\textwidth}
  \centering
  \includegraphics[width=.95\linewidth]{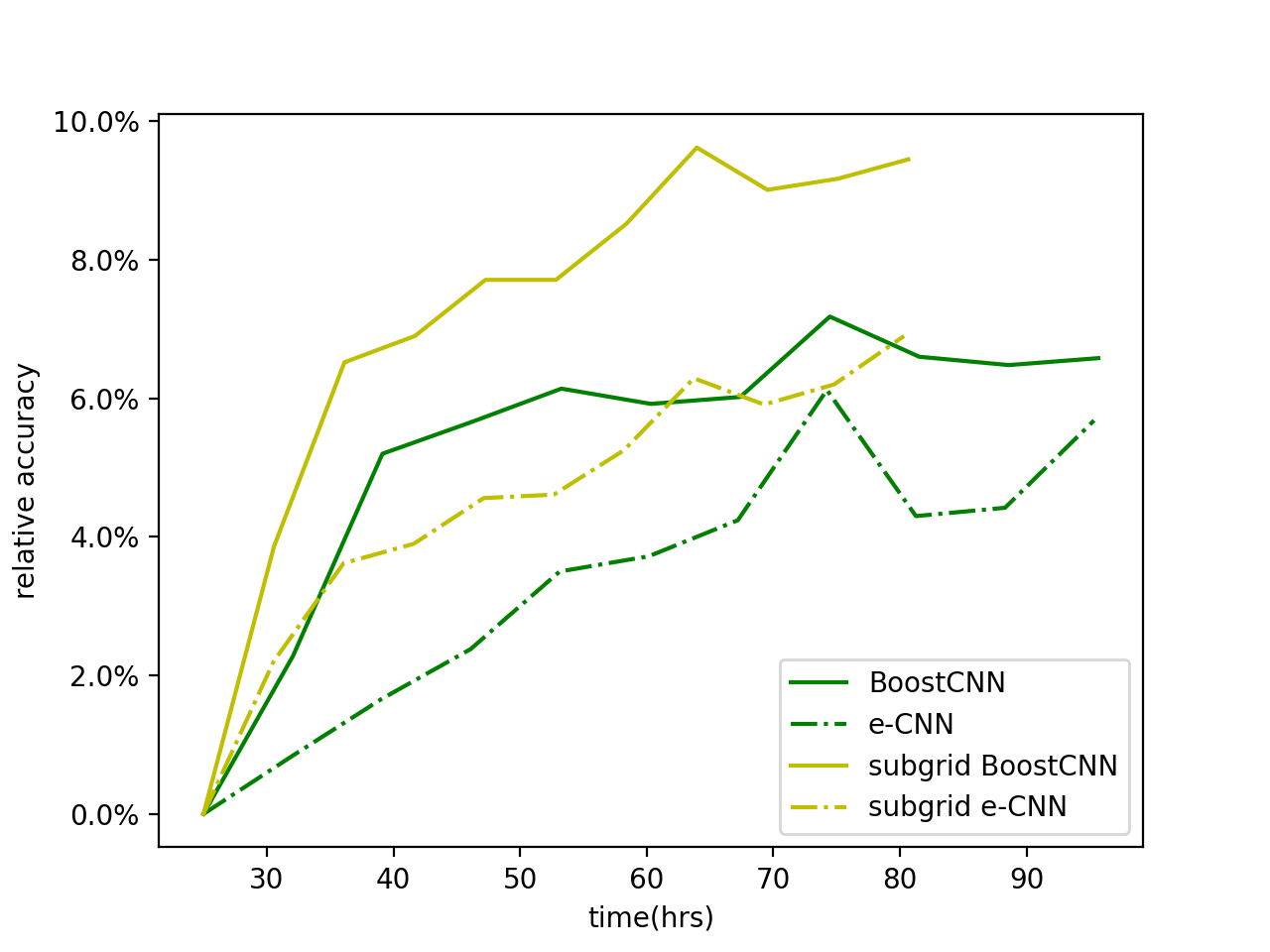}
  \captionof{figure}{ResNet-50 on ImageNetSub}
  \label{fig:50-image}
\end{minipage}%
\begin{minipage}[t]{.5\textwidth}
  \centering
  \includegraphics[width=.95\linewidth]{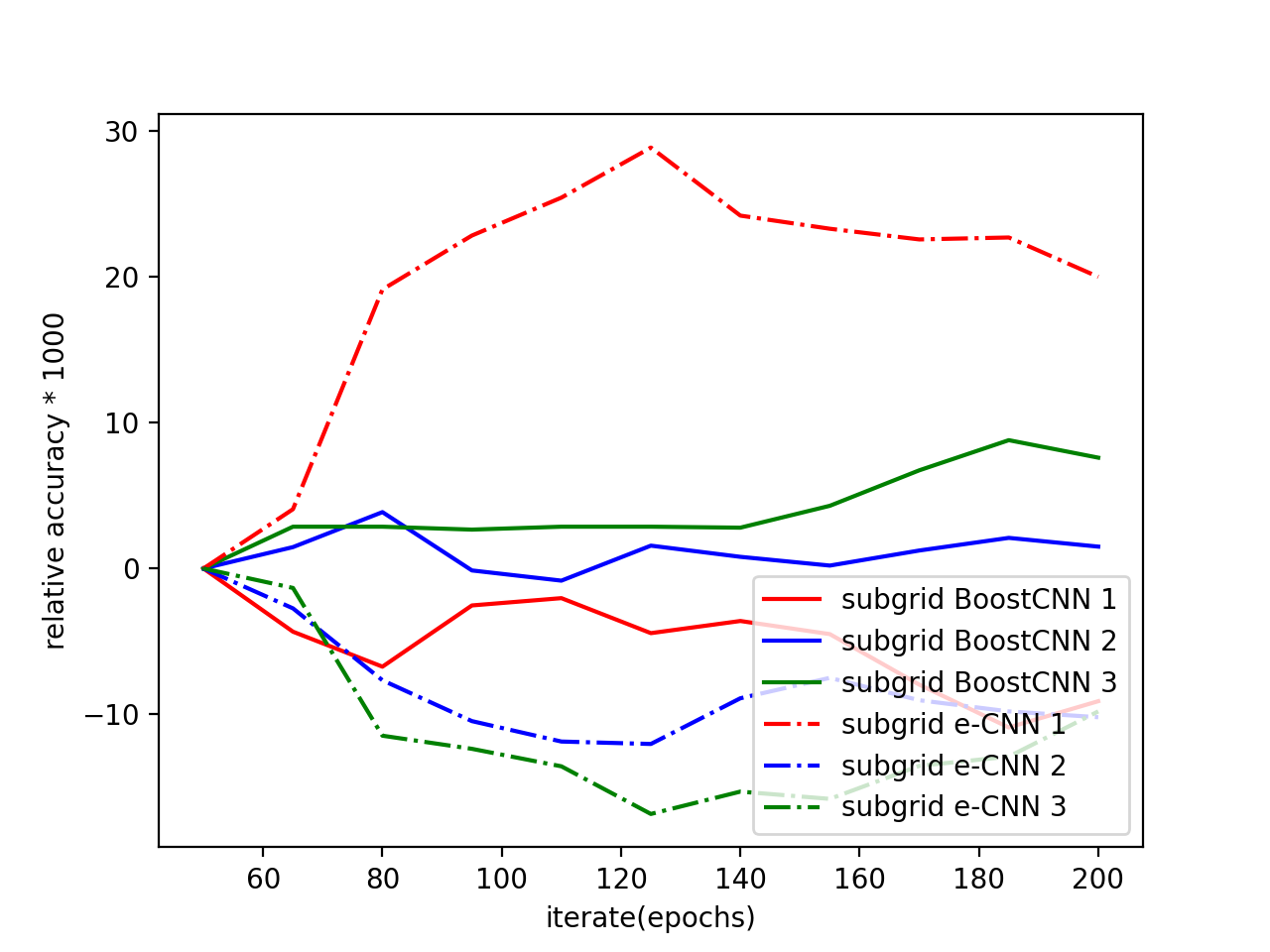}
  \captionof{figure}{Different Seeds}
  \label{fig:50-image-seed}
\end{minipage}
\begin{minipage}{.5\textwidth}
  \centering
  \includegraphics[width=.95\linewidth]{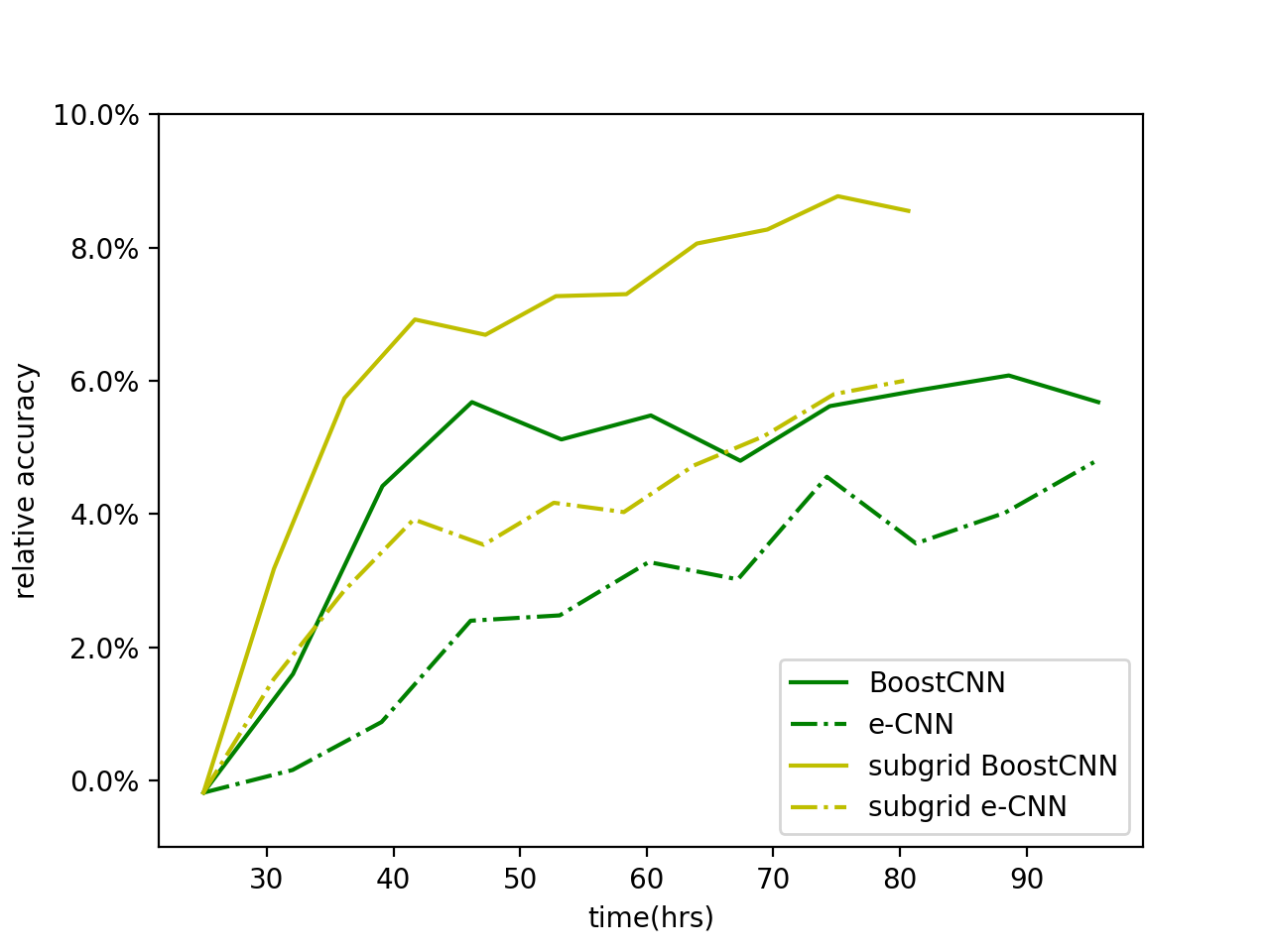}
  \captionof{figure}{ResNet-50 on ImageNetSub compared to ResNet-101}
  \label{fig:50-100 compare}
\end{minipage}%
\begin{minipage}{.5\textwidth}
  \centering
  \includegraphics[width=.95\linewidth]{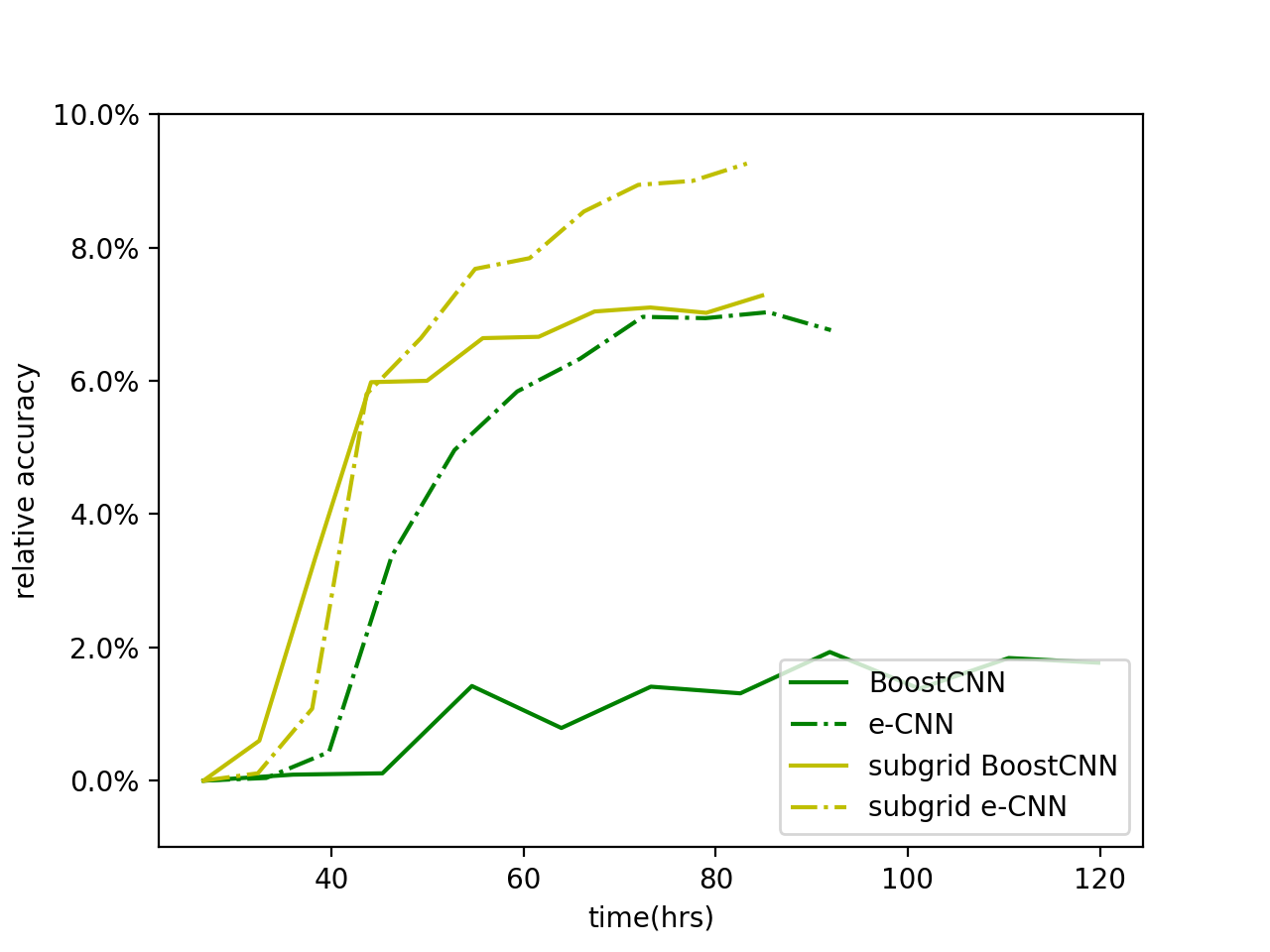}
  \captionof{figure}{ResNet-101 on ImageNetSub}
  \label{fig:101-image}
\end{minipage}
\end{figure}

We start by applying ResNet-18 as our weak learner for all different ensemble methods. Figures \ref{fig:18-cifar}, \ref{fig:18-svhn} and \ref{fig:18-image} compare the relative performances with respect to single ResNet-18 vs the running time. The solid lines in green and yellow show the relative performances of BoostCNN and subgrid BoostCNN, respectively, while the dotted lines in green and yellow represent the relative performances of e-CNN and subgrid e-CNN, respectively. As shown in these figures, taking the same amount of time, subgrid BoostCNN outperforms all of the remaining algorithms. Furthermore, we observe that subgrid BoostCNN outperforms BoostCNN, and subgrid e-CNN has the same behavior when compared with e-CNN. In conclusion, the subgrid technique improves the performance of the boosting algorithm. Moreover, Figures \ref{fig:18-cifar-seed}, \ref{fig:18-svhn-seed} and \ref{fig:18-image-seed} depict subgrid BoostCNN and subgrid e-CNN using three different seeds with respect to their averages. The solid and dotted lines in the same color represent the same seed used in corresponding subgrid BoostCNN and subgrid e-CNN. As the figures show, the solid lines are closer to each other than the dotted lines, which indicates that subgrid BoostCNN is more robust with respect to the variation of the seed when compared with subgrid e-CNN. Furthermore, the standard deviations of the accuracy generated by subgrid e-CNN and subgrid BoostCNN are shown in Table \ref{tb:seed}. The standard deviations of the accuracy generated by subgrid e-CNN are significant compared to those of subgrid BoostCNN, which in turn indicates that subgrid BoostCNN is less sensitive to the choice of the seed. Therefore, subgrid BoostCNN is more robust than subgrid e-CNN.
\begin{table}[H]
\centering
\begin{tabular}{|l|r|r|}
\hline
         & subgrid BoostCNN & subgrid e-CNN \\ \hline
CIFAR-10 & 0.478         & 2.519         \\ \hline
SVHN     & 0.385         & 0.891         \\ \hline
ImageNetSub & 2.489         & 7.915         \\ \hline
\end{tabular}
\caption{Standard deviation times $10^3$ of the accuracy results by different seeds}
\label{tb:seed}
\end{table}
Next, we evaluate relative performances of subgrid BoostCNN using ResNet-50 as the weak learner on CIFAR-10 and ImageNetSub datasets with respect to the single ResNet-50. We do not evaluate the relative performances on the SVHN dataset since the accuracy of the single ResNet-50 on the SVHN dataset is over 98$\%$. From Figures \ref{fig:50-cifar} and \ref{fig:50-image}, we also observe the benefits of the subgrid technique. Besides, Figures \ref{fig:50-cifar-seed} and \ref{fig:50-image-seed} confirm that subgrid BoostCNN is more stable than subgrid e-CNN since the solid series are closer to each other compared with the dotted series. Furthermore, we establish the relative performances of subgrid BoostCNN using ResNet-50 as the weak learner with respect to the single ResNet-101 in Figure \ref{fig:50-100 compare}. Although single ResNet-101 outperforms single ResNet-50, subgrid BoostCNN using ResNet-50 as the weak learner outperforms single ResNet-101 significantly in Figure \ref{fig:50-100 compare}, which indicates that subgrid BoostCNN with a simpler CNN is able to exhibit a better performance than a single deeper CNN. Lastly, we conduct experiments with ResNet-101 on the ImageNetSub dataset. From Figure \ref{fig:101-image}, we not only discover the superior behaviors of BoostCNN, e-CNN, subgrid BoostCNN and subgrid e-CNN over ResNet-101 as we expect, but also observe the benefit of the subgrid technique.

\subsection{Text}
In this section, we explore properties of the proposed Boost Transformer, subsequence Boost Transformer and importance-sampling-based Boost Transformer, and compare their performances with other methods on several text classification tasks. In the following experiments, the weak learner used is RoBERTa-based \cite{liu2019roberta} from the HuggingFace library with only word embeddings to be pre-trained weights. Using transformer based boosting algorithms, we train an ensemble of 6 transformers each with 5 epochs (these numbers yield good performance). In subsequence BoostTransformer, we pick the most important $80\%$ of the tokens in the vocabulary and reconstruct the dataset based on this new vocabulary. In importance-sampling-based BoostTransformer, the first flavor, in each iteration, we select $80\%$ of the samples based on the probability distribution in (\ref{importance sample distribution}) without further subsequence technique. In subsequence importance-sampling-based BoostTransformer, we first select $80\%$ of the samples based on the probability distribution in (\ref{importance sample distribution}), and then pick the most important $80\%$ of the tokens in the current vocabulary given by the selected $80\%$ samples, after that, we reconstruct the dataset based on this modified vocabulary. For comparison, we train the vanilla transformer and subsequence transformer, which randomly removes $20\%$ of the tokens and trains the network on the dataset for $30$ epochs. To train the model, we use AdamW \cite{Loshchilov2019DecoupledWD} with learning rate $10^{-5}$, weight decay $0.01$ and batch size $16$. We use linear learning rate decay with warmup ratio 0.06. 

We start by presenting the three public datasets used: IMDB \cite{Maas2011LearningWV}, Yelp polarity reviews and Amazon polarity reviews \cite{McAuley2013HiddenFA}. The IMDB dataset, which is for binary sentiment classification, contains a set of 25,000 highly polar movie reviews for training, and 25,000 for testing. The Yelp polarity reviews dataset, which is a subset of the dataset obtained from the Yelp Dataset Challenge in 2015, consists of $100,000$ training samples and $38,000$ testing samples. The classification task for this dataset is predicting a polarity label by considering stars 1 and 2 negative, and 3 and 4 positive for each review text. The last dataset we use is the Amazon polarity reviews dataset, which is a subset of the original Amazon reviews dataset from the Stanford Network Analysis Project (SNAP). Dealing with the same classification task as the Yelp polarity review dataset, the Amazon polarity reviews dataset contains $100,000$ training samples and $25,000$ testing samples. The subsampled datasets are standard, i.e. we did not create our own subsamples. Empirically we found that a weak learner with 6 heads and 6 layers achieves good robust performance.

\begin{figure}[H]
\begin{minipage}[t]{0.5\textwidth}
  \centering
  \includegraphics[width=.95\linewidth]{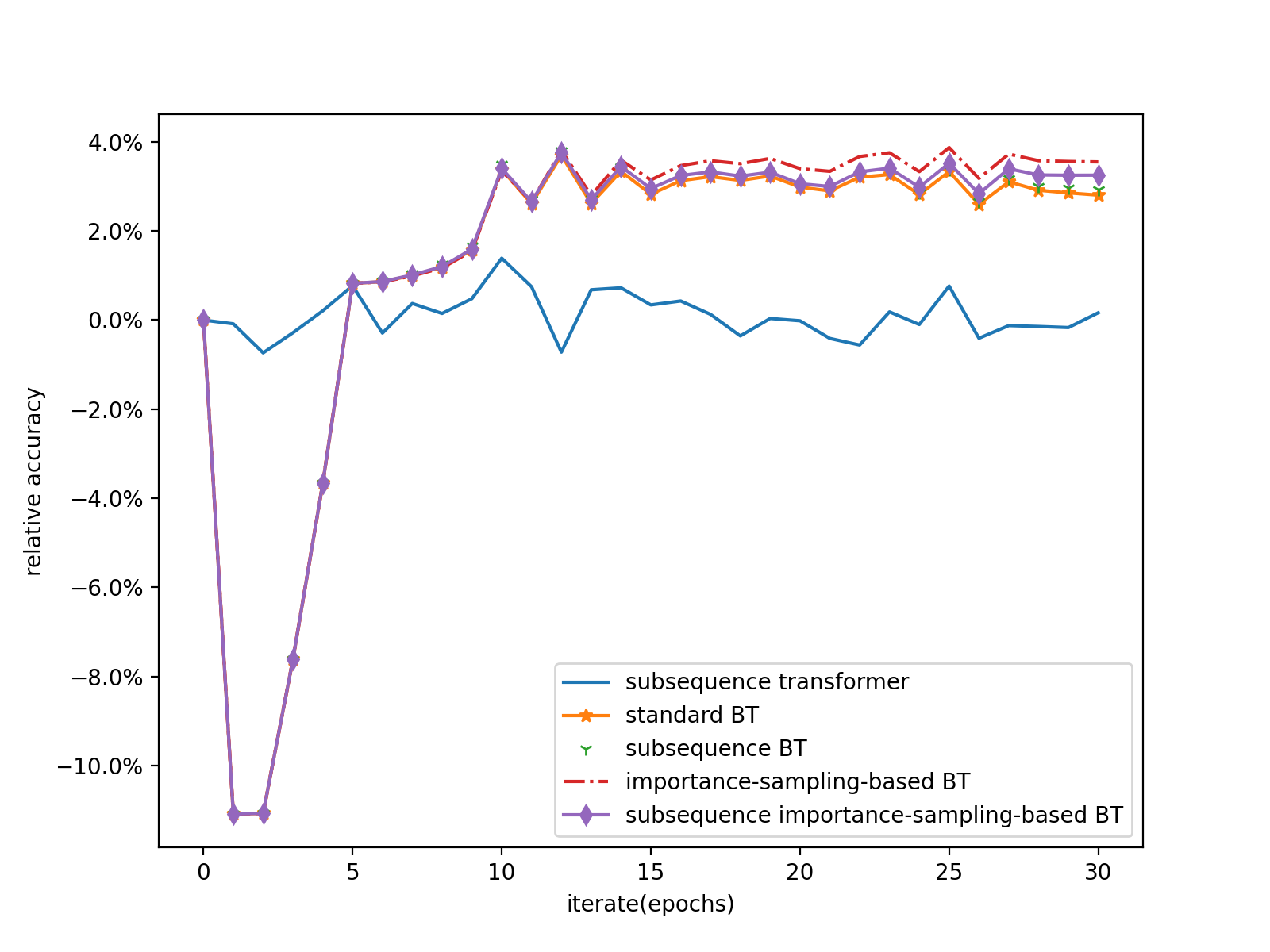}
  \captionof{figure}{Relative Accuracy on IMDB}
  \label{fig:imdb}
\end{minipage}%
\begin{minipage}[t]{0.5\textwidth}
  \centering
  \includegraphics[width=.95\linewidth]{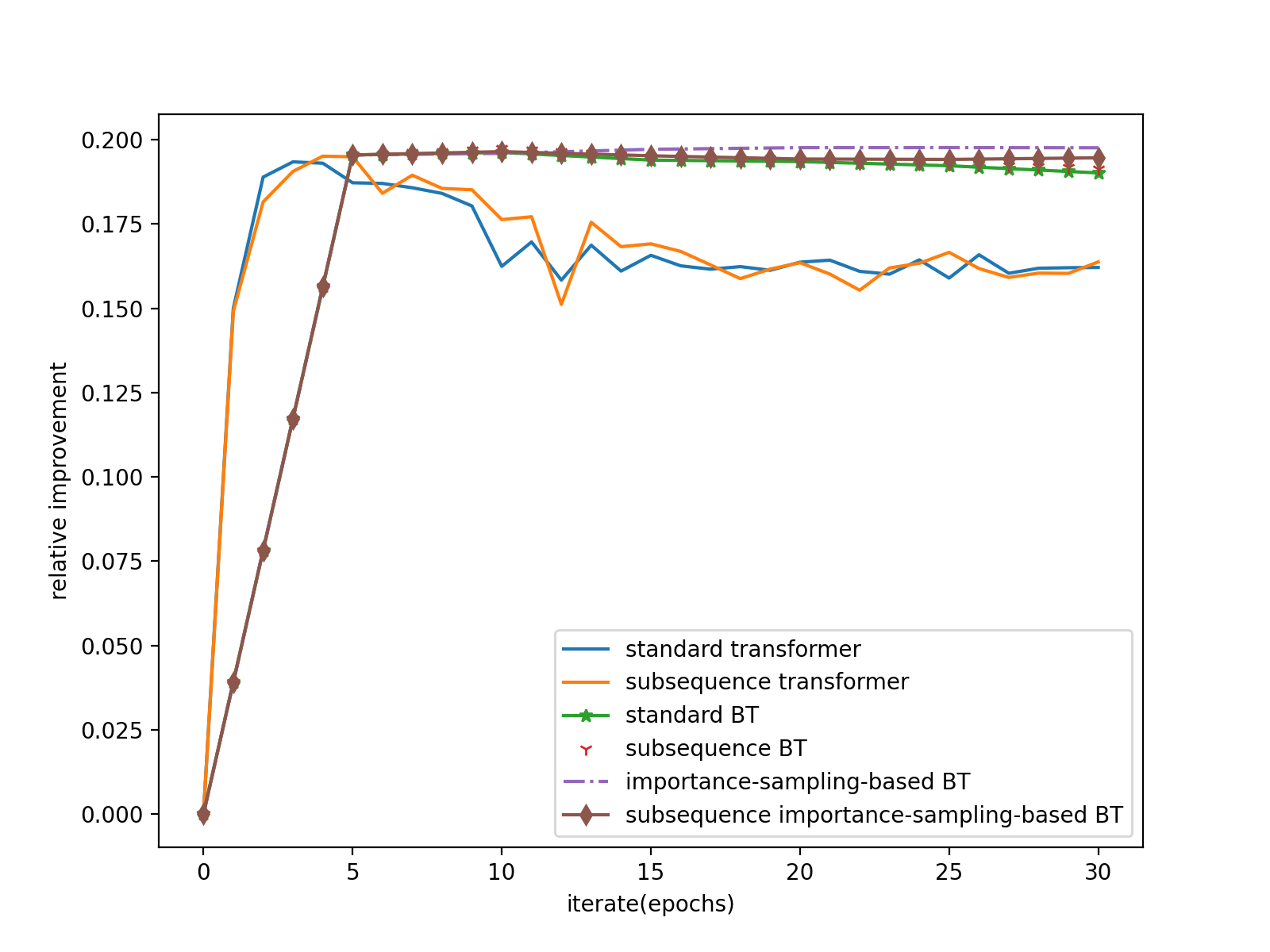}
  \captionof{figure}{Improvement on IMDB}
  \label{fig:imdb_imp}
\end{minipage}
\begin{minipage}[t]{0.5\textwidth}
  \centering
  \includegraphics[width=.95\linewidth]{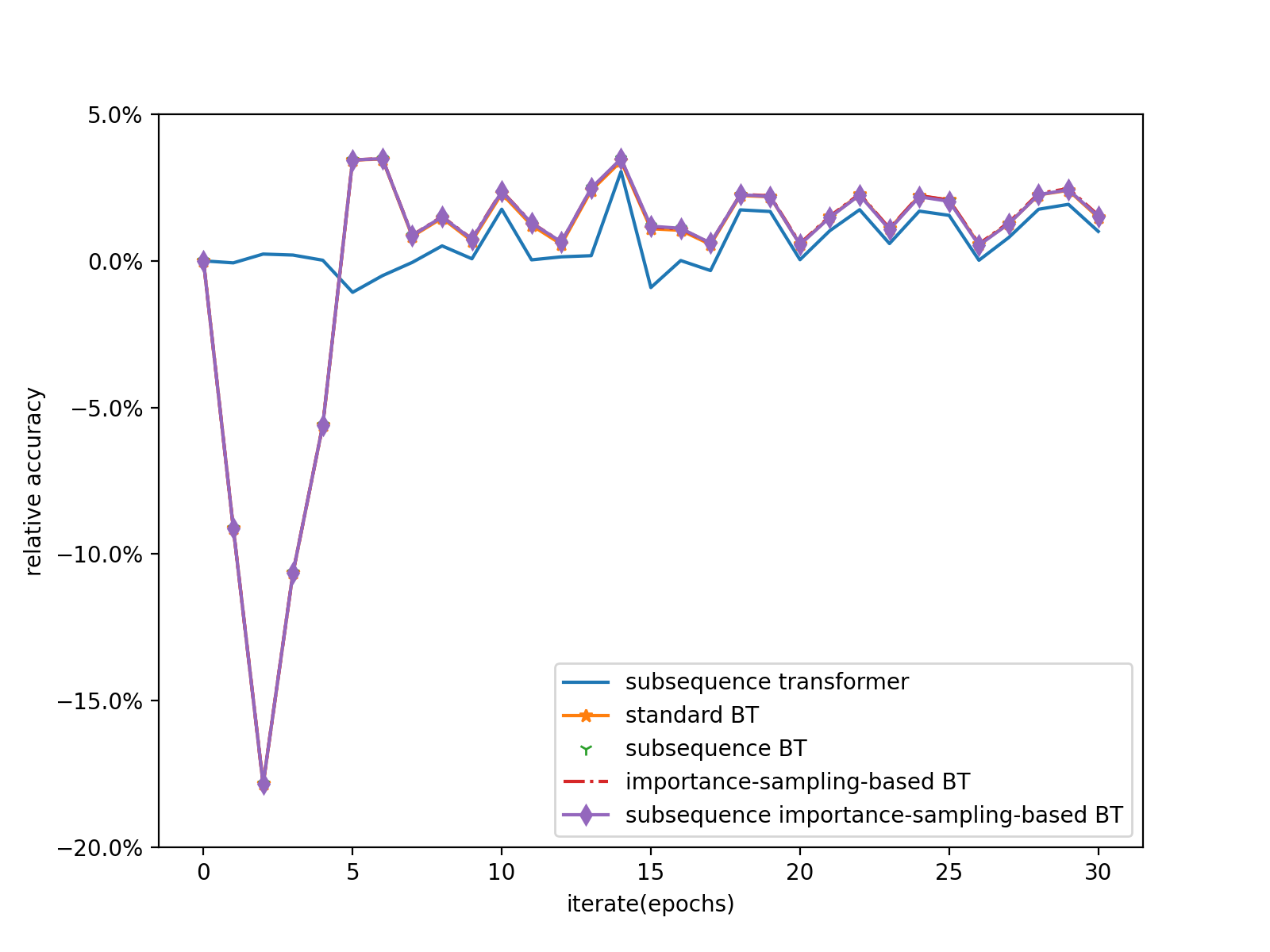}
  \captionof{figure}{Relative Accuracy on Yelp}
  \label{fig:yelp}
\end{minipage}%
\begin{minipage}[t]{0.5\textwidth}
  \centering
  \includegraphics[width=.95\linewidth]{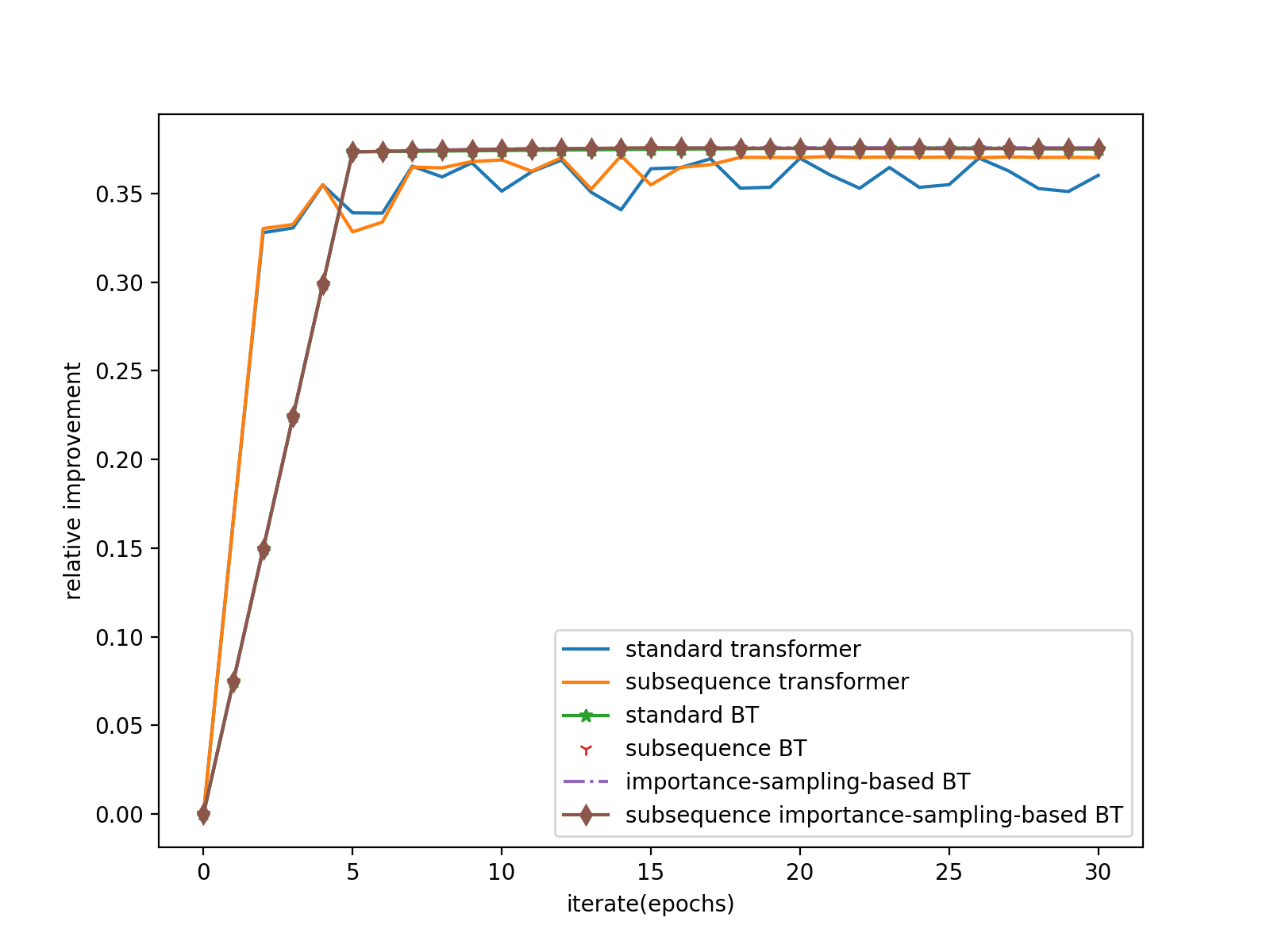}
  \captionof{figure}{Improvement on Yelp}
  \label{fig:yelp_imp}
\end{minipage}
\begin{minipage}[t]{.5\textwidth}
  \centering
  \includegraphics[width=.95\linewidth]{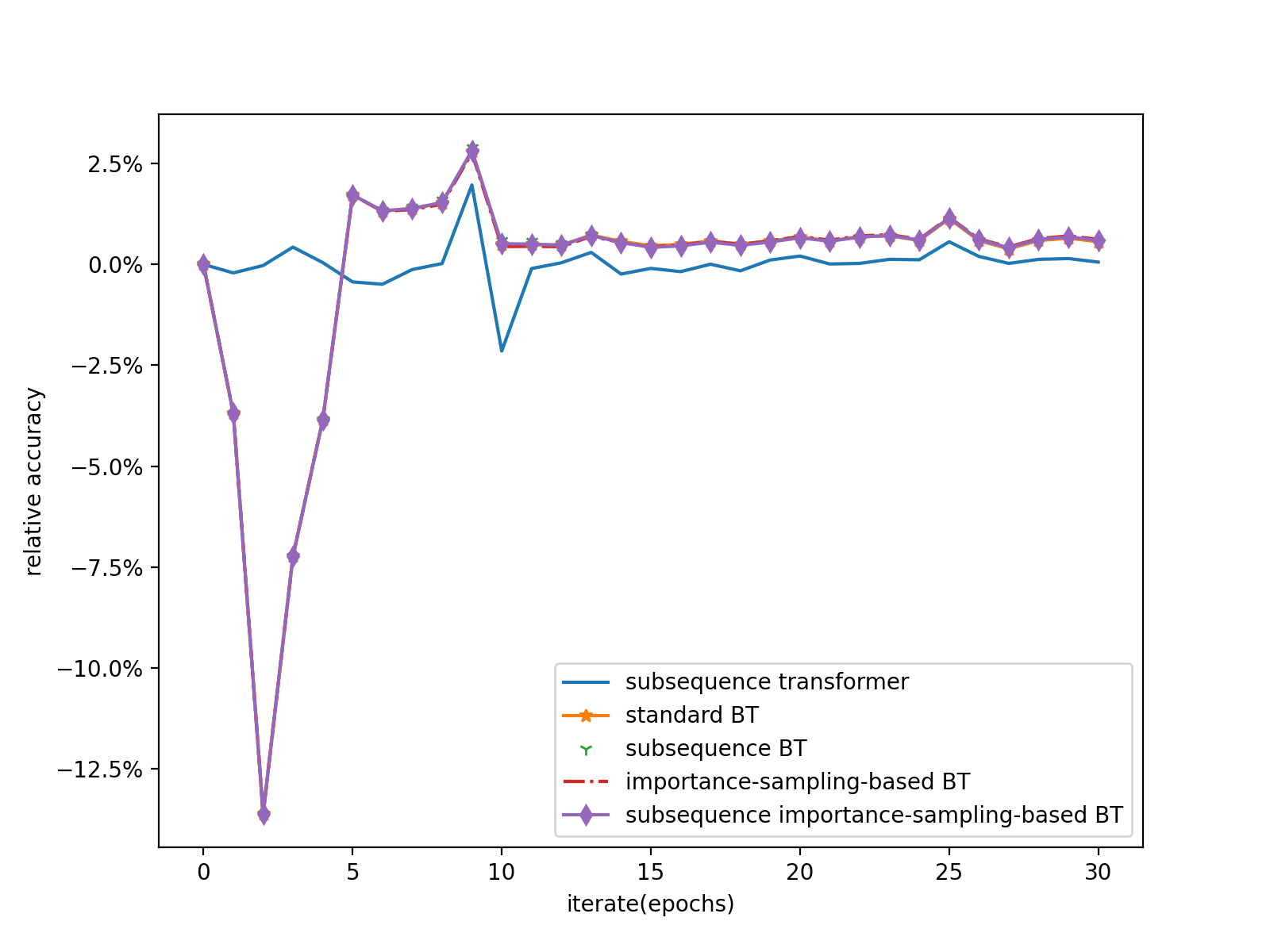}
  \captionof{figure}{Relative Accuracy on Amazon}
  \label{fig:amazon}
\end{minipage}%
\begin{minipage}[t]{.5\textwidth}
  \centering
  \includegraphics[width=.95\linewidth]{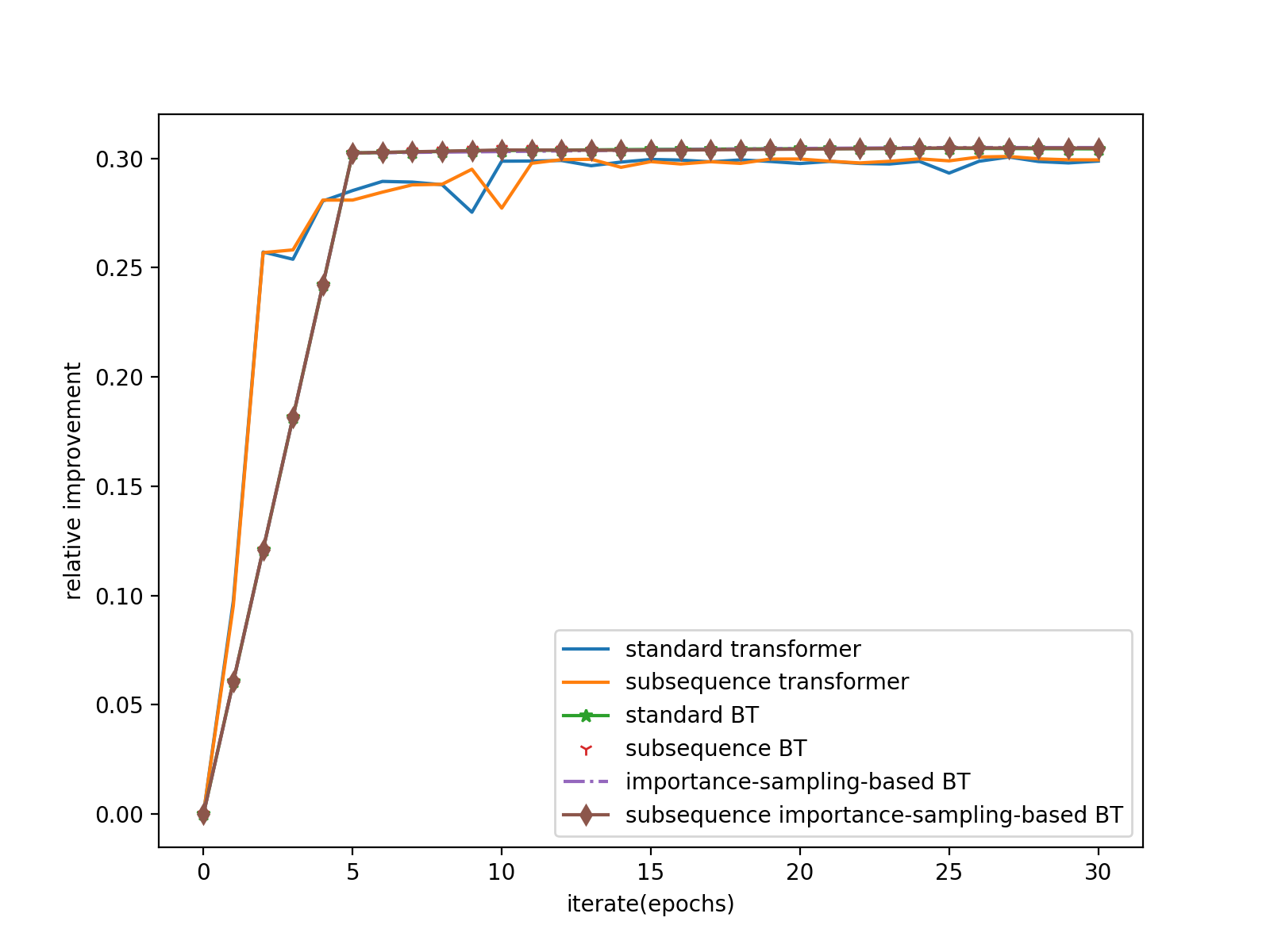}
  \captionof{figure}{Improvement on Amazon}
  \label{fig:amazon_imp}
\end{minipage}
\end{figure}

Given the architecture of the weak learner, we start by discussing experiments on IMDB. In Figure \ref{fig:imdb}, we compare the relative performances of the algorithms with respect to the vanilla transformer. As it shows, all versions of BoostTransformer do not perform as good as the standard transformer and subsequence transformer in the first few epochs. However, they catch up quickly and dominate the performance in the remaining training epochs. Even more, based on Figure \ref{fig:imdb_imp}, which represents each model's relative improvement with respect to its initial weights, all versions of BoostTransformer maintain their performances as the number of epochs increases, while the performances of the standard transformer and the subsequence transformer start decreasing and fluctuating dramatically after the first few epochs, which implies that all versions of BoostTransformer are more robust than the standard and subsequence transformer.  

Next, we evaluate relative performances with respect to the vanilla transformer on the Yelp and Amazon polarity review datasets. From Figures \ref{fig:yelp}-\ref{fig:amazon_imp}, we discover that the superior and more robust behavior of boosting algorithms over transformer is vigorous. 

\begin{figure}[H]
\begin{minipage}{.9\textwidth}
  \centering
  \includegraphics[width=.95\linewidth]{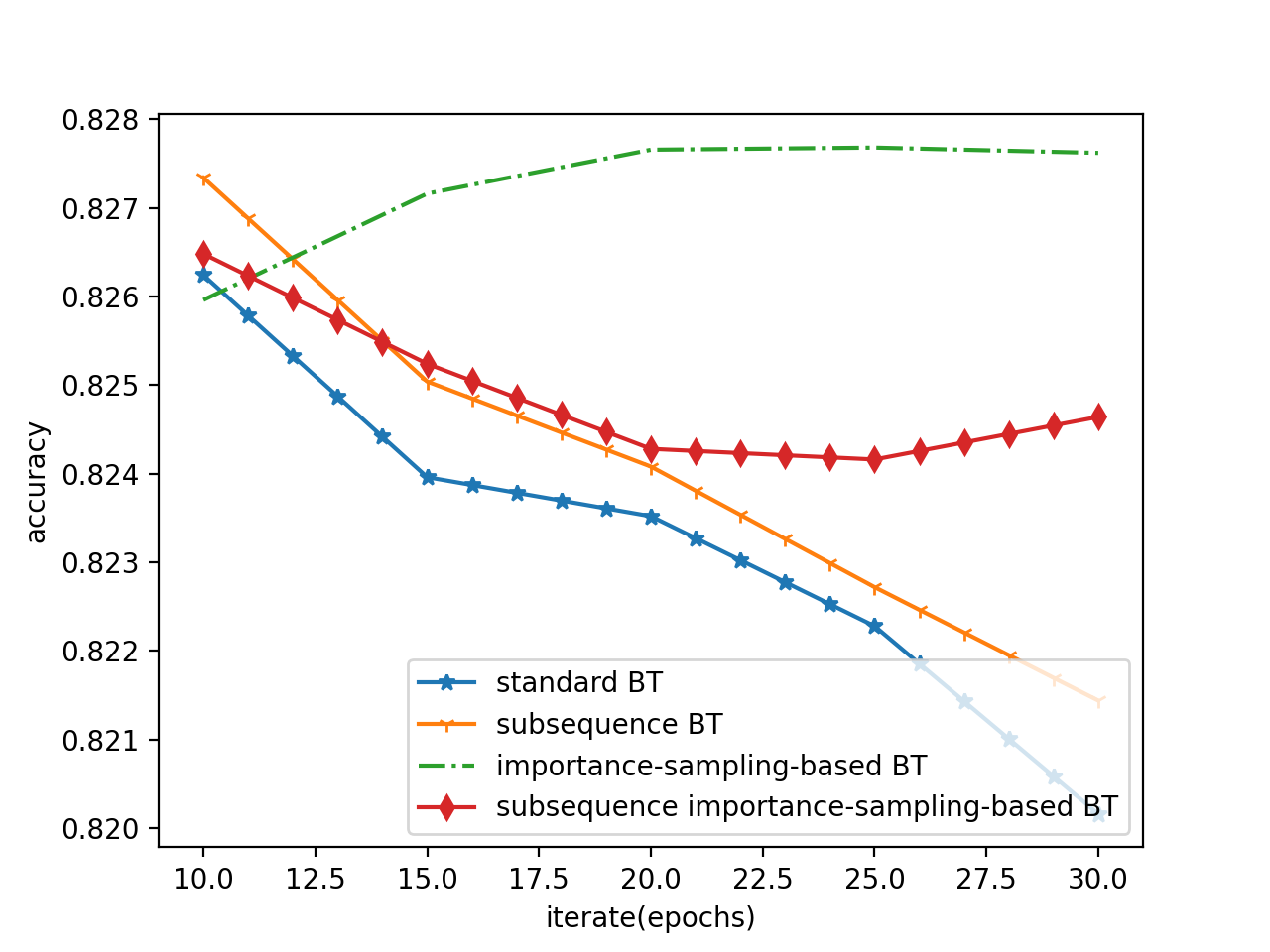}
  \captionof{figure}{IMDB}
  \label{fig:imdb_zoom}
\end{minipage}
\begin{minipage}{.9\textwidth}
  \centering
  \includegraphics[width=.95\linewidth]{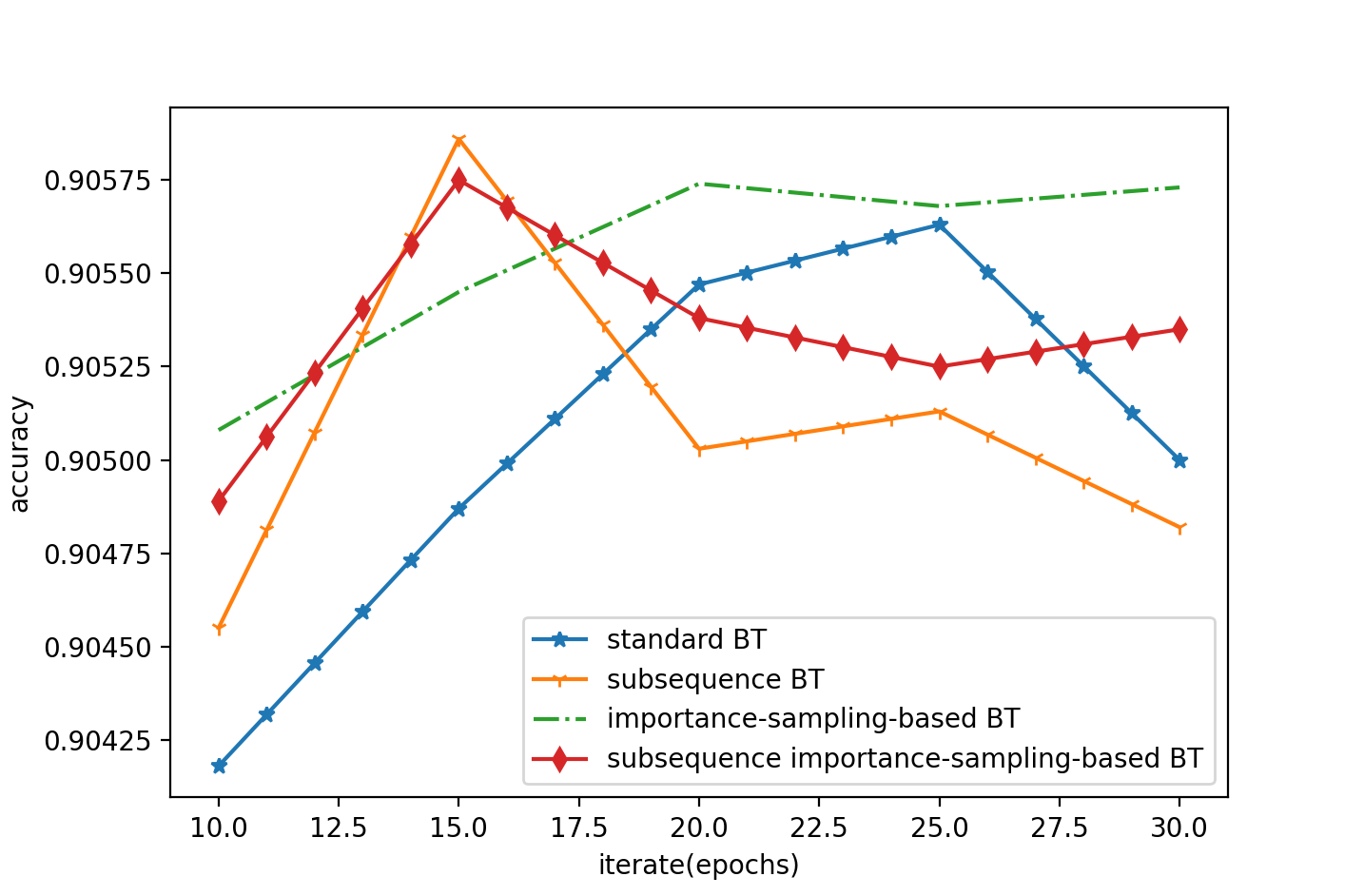}
  \captionof{figure}{Yelp}
  \label{fig:yelp_zoom}
\end{minipage}
\end{figure}

\begin{figure}
\begin{minipage}{.9\textwidth}
  \centering
  \includegraphics[width=.95\linewidth]{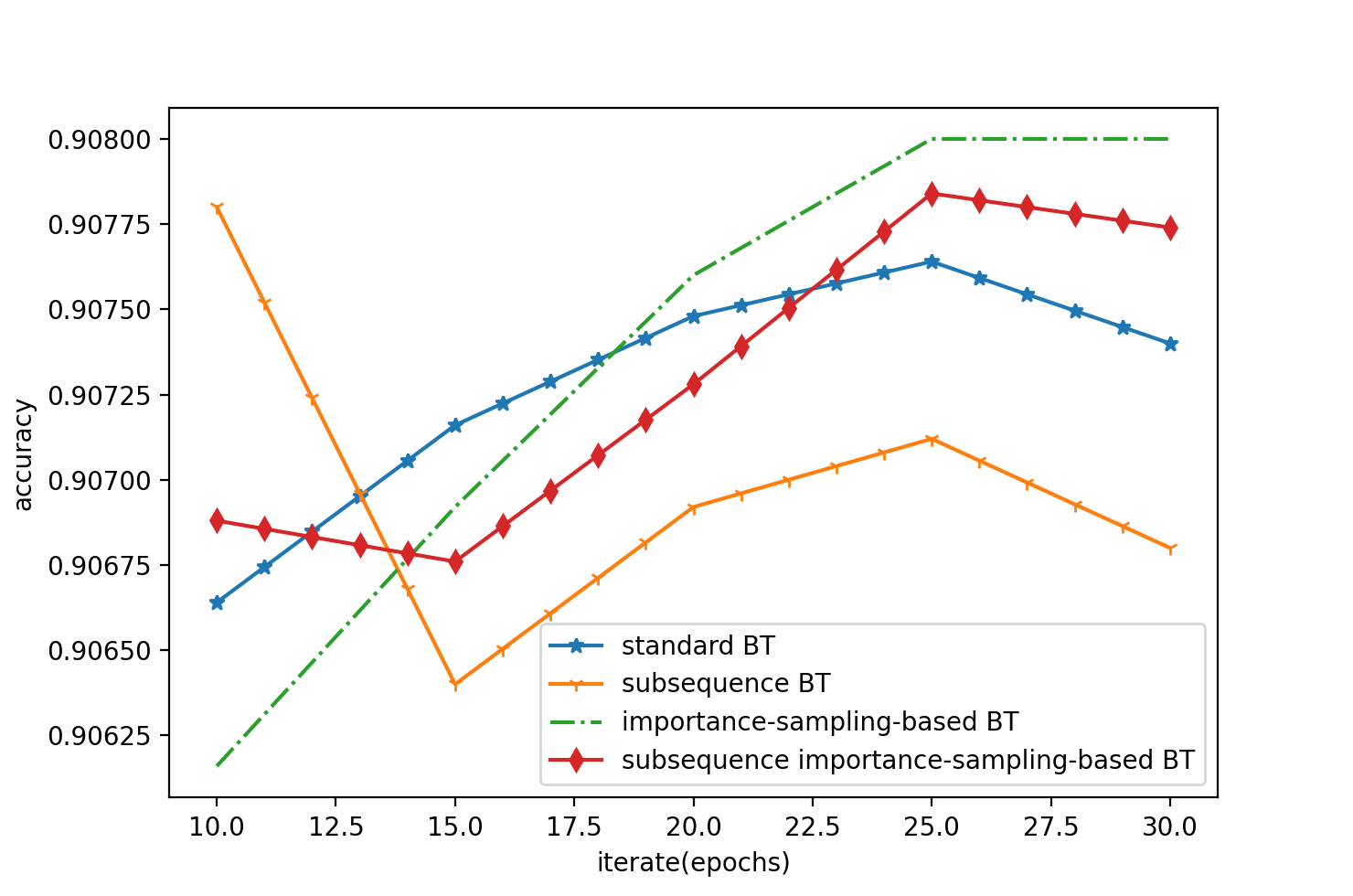}
  \captionof{figure}{Amazon}
  \label{fig:amazon_zoom}
\end{minipage}
\end{figure}

Furthermore, we zoom in on the performances at iterates larger than 2. In Figures \ref{fig:imdb_zoom}-\ref{fig:amazon_zoom}, compared with the standard BoostTransformer, we observe that the subsequence BoostTransformer, importance-sampling-based BoostTransformer and subsequence importance-sampling-based BoostTransformer demonstrate a superior performance. Therefore, we conclude that the subsequence and importance sampling techniques are beneficial for the boosting algorithms. Moreover, we observe that the importance-sampling-based BoostTransformer gradually improves its performance and maintains its performance later on, while the subsequence BoostTransformer hits its best accuracy in early epochs and then starts fluctuating and decaying.  The gap between the importance-sampling-based BoostTransformer and the subsequence BoostTransformer is more significant on the IMDB dataset, which has a much smaller size than the Yelp and Amazon polarity review datasets. For the subsequence importance-sampling-based BoostTransformer, compared to the subsequence BoostTransformer, although the subsequence importance-sampling-based BoostTransformer does not fluctuate and decrease as much as the subsequence BoostTransformer, which is more obvious in a small dataset (i.e. the IMDB dataset), its best accuracy is lower than that of the subsequence BoostTransformer, which is more obvious in larger datasets (i.e. the Yelp and Amazon datasets). On the other hand, compared to the importance-sampling-based BoostTransformer, although the subsequence importance-sampling-based BoostTransformer obtains its best accuracy earlier than the importance-sampling-based BoostTransformer,  its overall performance fluctuates while the importance-sampling-based BoostTransformer keeps increasing and maintains its high-quality performance in all of the datasets, which implies that the subsequence importance-sampling-based BoostTransformer is less stable than the importance-sampling-based BoostTransformer. In conclusion, the subsequence BoostTransformer fits well for datasets with enough samples and the importance-sampling-based BoostTransformer is more suitable for datasets with a limited number of samples.
Table \ref{tb:timeTrans} illustrates the running time of each algorithm on the different datasets in minutes. As we see in the table, the subsequence technique not only improves the performance of the boosting algorithms but also reduces the running time. Furthermore, the importance sampling technique reduces the running time significantly without hurting the performance.
\begin{table}[H]
\centering
\begin{tabular}{|l|r|r|r|r|r|r|}
\hline
{\color[HTML]{333333} }               & \multicolumn{1}{c|}{{\color[HTML]{333333} Trans.}}  &\multicolumn{1}{l|}{\begin{tabular}[c]{@{}l@{}}subsequence \\ Trans.\end{tabular}}  & \multicolumn{1}{l|}{BT} & \multicolumn{1}{l|}{\begin{tabular}[c]{@{}l@{}}subsequence \\ BT\end{tabular}} & \multicolumn{1}{l|}{\begin{tabular}[c]{@{}l@{}}importance-sampling\\-based BT\end{tabular}} &\multicolumn{1}{l|}{\begin{tabular}[c]{@{}l@{}}subsequence\\importance-sampling\\-based BT\end{tabular}}\\ \hline
IMDB&21 & 14 & 23 & 17 & 13& 12\\  \hline
Yelp &  76& 52 & 84 & 66 & 52& 46\\ \hline
 Amazon &75 & 53 & 79 & 58& 46& 41\\ \hline
\end{tabular}
\caption{Running time for different algorithms}\label{tb:timeTrans}
\end{table}
In conclusion, a subsequence transformer is a good choice if the running time cost is the most important concern, however, if accuracy performance is as crucial as the running time, then the subsequence BoostTransformer is the go-to option since it requires a slight increase in the running time but provides superior and more robust performance when compared to the subsequence transformer. In addition, if a dataset has a limited number of samples, i.e., it is easy to cause overfitting, then the importance-sampling-based BoostTransformer can outperform.

\newpage
\bibliographystyle{plain}
\bibliography{references}

\newpage
\section{Appendix}
\subsection*{A \tab Proof of Theorem \hyperref[thm:1]{1}\label{pf:thm1}}
\begin{proof}
Given a probability distribution $P$ for dataset $(x_i,z_i)$, by assumption, the stochastic gradient of the loss function is unbiased, i.e.
\begin{align}
\label{eq:unbiased grad}
\mathbb{E}_{P}\left[\frac{\partial \bar{L}_i(z_i,f(x_i))}{\partial f} \right]=\frac{\partial \mathcal{R}}{\partial f},
\end{align}
with
\begin{align*}
  \mathcal{R}[f] = \frac{1}{\left|\mathcal{D}\right|}\sum_{(x_i,z_i)\in\mathcal{D}}L(z_i,f(x_i))=\mathbb{E}_{P}\left[\bar{L}_i(z_i,f(x_i)) \right]
\end{align*}
and
\begin{align*}
    \bar{L}_i(z_i,f(x_i))=\frac{1}{\left|\mathcal{D}\right|P(I=i)}L(z_i,f(x_i)).
\end{align*}

At iterate $t$, in importance-sampling-based Boosting algorithms, given probability distribution $P_t$ and $P_t^i = P_t(I=i)$, the current gradient given a subset $\mathcal{I}^t$ of samples is
\begin{align}
\label{eq:grad_explain}
    \bar{g}_t^{\mathcal{I}^t}&=\frac{1}{\left|\mathcal{I}^t\right|}\sum_{(x_i,z_i)\in\mathcal{I}^t}\left.\frac{\partial \bar{L}_i(z_i,f_{t-1}(x_i)+\epsilon g(x_i))}{\partial g}\right|_{\epsilon =0}\nonumber\\
    &=\frac{1}{\left|\mathcal{I}^t\right|}\sum_{(x_i,z_i)\in\mathcal{I}^t}\frac{1}{\left|\mathcal{D}\right|P^i_t}\left.\frac{\partial L_i(z_i,f_{t-1}(x_i)+\epsilon g(x_i))}{\partial g}\right|_{\epsilon =0}\nonumber\\
    &= \frac{1}{\left|\mathcal{I}^t\right|}\sum_{(x_i,z_i)\in\mathcal{I}^t}\frac{1}{\left|\mathcal{D}\right|P^i_t} g_t^i=\frac{1}{\left|\mathcal{I}^t\right|}\sum_{k=1}^{\left|\mathcal{I}^t\right|}G_k,
\end{align}
where $g_t^i = \left.\frac{\partial L_i(z_i,f_{t-1}(x_i)+\epsilon g(x_i))}{\partial g}\right|_{\epsilon =0}$ and $G_k$ is the random variable corresponding to sample $k$. Note that 
\begin{align}
    g_t=\frac{\partial \mathcal{R}[f_{t-1};g]}{\partial g}=\left.\frac{\partial \mathcal{R}[f_{t-1}+\epsilon g]}{\partial g}\right|_{\epsilon =0}
\end{align}
and
\begin{align}
\label{eq:importance grad}
    \mathbb{E}_{P_t}(G_t)=\mathbb{E}_{P_t}\left[\bar{g}_t^{\mathcal{I}^t} \right]=g_t,
\end{align}
due to the unbiased gradient in (\ref{eq:unbiased grad}). Given $\bar{g}_{t}^{\mathcal{I}^{t}}$ computed on a subset $\mathcal{I}^{t}$ with probability distribution $P_{t}$, we consider
\begin{align}
    \mathbb{E}_{P_{t}}\left[ \Delta^{(t)} \right]&=\left\|f_{t-1}-f^*\right\|^2-\mathbb{E}_{P_{t}}\left[\left.\left\| f_{t}-f^*\right\|^2 \right|\mathcal{F}^{t-1}\right]\nonumber\\
    &=\left\|f_{t-1}-f^*\right\|^2-\mathbb{E}_{P_{t}}\left[\left.\left\| f_{t-1}+\alpha_{t}\bar{g}_{t}^{\mathcal{I}^{t}}-f^*\right\|^2 \right|\mathcal{F}^{t-1}\right]\nonumber\\
    &=-2\alpha_{t}\left\langle f_{t-1}-f^*, \mathbb{E}_{P_{t}}\left[\left.\bar{g}_{t}^{\mathcal{I}^{t}}\right|\mathcal{F}^{t-1}\right]\right\rangle -\alpha_{t}^2 \mathbb{E}_{P_{t}}\left[\left.\left\| \bar{g}_{t}^{\mathcal{I}^{t}}\right\|^2 \right|\mathcal{F}^{t-1}\right].\label{eq:diff1}
\end{align}
By inserting (\ref{eq:importance grad}) into (\ref{eq:diff1}), we have
\begin{align}
\label{eq:diff2}
     \mathbb{E}_{P_{t}}\left[ \Delta^{(t)} \right]=-2\alpha_{t}\left\langle f_{t-1}-f^*, g_{t}\right\rangle -\alpha_{t}^2 \mathbb{E}_{P_{t}}\left[\left.\left\| \bar{g}_{t}^{\mathcal{I}^{t}}\right\|^2 \right|\mathcal{F}^{t-1}\right].
\end{align}
Thus, maximizing $\mathbb{E}_{P_{t}}\left[ \Delta^{(t)} \right]$ is equivalent to minimizing the variance of the gradient, i.e. $\mathbb{E}_{P_{t}}\left[\left.\left\| \bar{g}_{t}^{\mathcal{I}^{t}}\right\|^2 \right|\mathcal{F}^{t-1}\right]$. Consequently, consider
\begin{align}
\label{eq:min_imp}
    \mathbb{E}_{P_{t}}\left[\left.\left\| \bar{g}_{t}^{\mathcal{I}^{t}}\right\|^2 \right|\mathcal{F}^{t-1}\right] &=\mathbb{E}_{P_{t}}\left[\left.\left\| \bar{g}_{t}^{\mathcal{I}^{t}}-g_t+g_t\right\|^2 \right|\mathcal{F}^{t-1}\right]\nonumber\\
    &=\mathbb{E}_{P_{t}}\left[\left.\left\| \bar{g}_{t}^{\mathcal{I}^{t}}-g_t\right\|^2 \right|\mathcal{F}^{t-1}\right]+
    \mathbb{E}_{P_{t}}\left[\left. 2<\bar{g}_{t}^{\mathcal{I}^{t}}-g_t,g_t> \right|\mathcal{F}^{t-1}\right]+\left\|g_t\right\|^2    \nonumber\\
    &=\mathbb{E}_{P_{t}}\left[\left.\left\| \bar{g}_{t}^{\mathcal{I}^{t}}-g_t\right\|^2 \right|\mathcal{F}^{t-1}\right]+2<\mathbb{E}_{P_{t}}\left[\left.\bar{g}_{t}^{\mathcal{I}^{t}}\right|\mathcal{F}^{t-1}\right]-g_t,g_t> +\left\|g_t\right\|^2    \nonumber\\
    &=\mathbb{E}_{P_{t}}\left[\left.\left\| \bar{g}_{t}^{\mathcal{I}^{t}}-g_t\right\|^2 \right|\mathcal{F}^{t-1}\right]+\left\|g_t\right\|^2,
\end{align}
where the last equality holds due to (\ref{eq:importance grad}). Continuing, we have
\begin{align}
    &\mathbb{E}_{P_{t}}\left[\left.\left\| \bar{g}_{t}^{\mathcal{I}^{t}}-g_t\right\|^2 \right|\mathcal{F}^{t-1}\right]= \mathbb{E}_{P_{t}}\left[\left.\left\| \frac{1}{\left|\mathcal{I}^{t}\right|}\sum_{i\in\mathcal{I}^{t}}\left(\frac{1}{\left|\mathcal{D}\right|P_t^i} g_{t}^i-g_t\right)\right\|^2 \right|\mathcal{F}^{t-1}\right]\nonumber\\
    &=\frac{1}{\left|\mathcal{I}^{t}\right|^2}\mathbb{E}_{P_{t}}\left[\left.\left\| \sum_{i\in\mathcal{I}^{t}}\left(\frac{1}{\left|\mathcal{D}\right|P_t^i} g_{t}^i-g_t\right)
    \right\|^2 \right|\mathcal{F}^{t-1}\right]\nonumber\\
    &=\frac{1}{\left|\mathcal{I}^{t}\right|^2}\mathbb{E}_{P_{t}}\left[\left.\sum_{i\in\mathcal{I}^{t}}\left\| \frac{1}{\left|\mathcal{D}\right|P_t^i} g_{t}^i-g_t\right\|^2+\sum_{(i,j)\in\mathcal{I}^{t},i\neq j}<\frac{1}{\left|\mathcal{D}\right|P_t^i} g_{t}^i-g_t, \frac{1}{\left|\mathcal{D}\right|P_t^j} g_{t}^j-g_t> \right|\mathcal{F}^{t-1}\right]\nonumber\\
    &=\frac{1}{\left|\mathcal{I}^{t}\right|^2}\left(\mathbb{E}_{P_{t}}\left[\left.\sum_{i\in\mathcal{I}^{t}}\left\| \frac{1}{\left|\mathcal{D}\right|P_t^i} g_{t}^i-g_t\right\|^2\right|\mathcal{F}^{t-1}\right]+\mathbb{E}_{P_{t}}\left[\left. \sum_{(i,j)\in\mathcal{I}^{t},i\neq j}<\frac{1}{\left|\mathcal{D}\right|P_t^i} g_{t}^i-g_t, \frac{1}{\left|\mathcal{D}\right|P_t^j} g_{t}^j-g_t> \right|\mathcal{F}^{t-1}\right]\right)\nonumber\\
    &=\frac{|\mathcal{I}^{t}|}{\left|\mathcal{I}^{t}\right|^2}\mathbb{E}_{P_{t}}\left[\left.\left\| G_1-g_t\right\|^2\right|\mathcal{F}^{t-1}\right]+\frac{2}{\left|\mathcal{I}^{t}\right|^2}\binom{\left|\mathcal{I}^{t}\right|}{2} <\mathbb{E}_{P_{t}}\left[\left.G_1-g_t\right|\mathcal{F}^{t-1}\right], \mathbb{E}_{P_{t}}\left[\left.G_2-g_t\right|\mathcal{F}^{t-1}\right]> \nonumber\\
    &= \frac{1}{\left|\mathcal{I}^{t}\right|}\mathbb{E}_{P_{t}}\left[\left.\left\| G_1-g_t
    \right\|^2 \right|\mathcal{F}^{t-1}\right]\nonumber\\
    &=\frac{1}{\left|\mathcal{I}^{t}\right|}\left(\mathbb{E}_{P_{t}}\left[\left.\left\| G_1
    \right\|^2 \right|\mathcal{F}^{t-1}\right]-\left\|g_t\right\|^2\right).
    \label{eq:final_min}
\end{align}
The fifth equality holds since $G_i$ and $G_j$ are independent, moreover, the seventh equality is valid due to (\ref{eq:importance grad}). Inserting (\ref{eq:final_min}) into (\ref{eq:min_imp}) yields
\begin{align}
    \mathbb{E}_{P_{t}}\left[\left.\left\| \bar{g}_{t}^{\mathcal{I}^{t}}\right\|^2 \right|\mathcal{F}^{t-1}\right] = \frac{1}{\left|\mathcal{I}^{t}\right|\left|\mathcal{D}\right|^2}\sum_{(x_i,z_i)\in\mathcal{D}}\frac{1}{P_t^i}\left\| g_{t}^i\right\|^2 -\frac{1}{\left|\mathcal{I}^{t}\right|}\left\|g_t\right\|^2 + \left\|g_t\right\|^2.
\label{eq:final_goal_imp}
\end{align}
As (\ref{eq:final_goal_imp}) shows, maximizing $\mathbb{E}_{P_{t}}\left[ \Delta^{(t)} \right]$ is equivalent to minimizing $\frac{1}{P_t^i}\left\| g_{t}^i\right\|^2$. By using the Jensen's inequality, it follows that
\begin{align}
    \sum_{(x_i,z_i)\in\mathcal{D}}\frac{1}{P_t^i}\left\| g_{t}^i\right\|^2=\sum_{(x_i,z_i)\in\mathcal{D}}P_t^i\left( \frac{\left\|g_{t}^i\right\|}{P_t^i}\right)^2\geq \left(\sum_{(x_i,z_i)\in\mathcal{D}} \left\|g_{t}^i\right\|\right)^2,
\end{align}
and the equality holds when $P_t^i=\left\|g_{t}^i\right\|/\sum_{(x_j,z_j)\in\mathcal{D}}\left\|g_{t}^j\right\|$. Note that $g_{t}^i=\left.\frac{\partial L(z_i,f_{t-1}(x_i)+\epsilon g(x_i)}{\partial g}\right|_{\epsilon =0}$ is proportional to the boosting weights $w_{t}(x_i,z_i)$ of sample $(x_i,z_i)$ as stated in (\ref{eq:grad-weight}), therefore, the claim in (\ref{importance sample distribution}) follows.
\end{proof}

\end{document}